\documentclass[sigconf]{acmart}

\RequirePackage[varqu]{zi4}
\RequirePackage[libertine]{newtxmath}

\usepackage{booktabs} 

\usepackage{amsmath}
\usepackage[caption=false,font=footnotesize]{subfig}
\usepackage{url}
\usepackage{hyperref}

\usepackage{graphicx}
\usepackage{braket}
\usepackage{amsfonts}
\usepackage{amssymb}
\usepackage{amsthm}
\usepackage{color}
\usepackage{mydef}
\usepackage{cleveref}
\crefformat{equation}{(#2#1#3)}
\crefrangeformat{equation}{(#3#1#4) to~(#5#2#6)}
\crefmultiformat{equation}{(#2#1#3)}%
{ and~(#2#1#3)}{, (#2#1#3)}{ and~(#2#1#3)}

\copyrightyear{2018}
\acmYear{2018}
\setcopyright{acmcopyright}
\acmConference[GECCO '18]{Genetic and Evolutionary Computation Conference}{July 15--19, 2018}{Kyoto, Japan} \acmBooktitle{GECCO '18: Genetic and Evolutionary Computation Conference, July 15--19, 2018, Kyoto, Japan} \acmPrice{15.00} \acmDOI{10.1145/3205455.3205613} \acmISBN{978-1-4503-5618-3/18/07}

\begin{document}
\title{Data-Driven Analysis of Pareto Set Topology}

\author{Naoki Hamada}
\affiliation{%
  \institution{Fujitsu Laboratories Ltd.}
  \streetaddress{4-1-1 Kamikodanaka, Nakahara-ku}
  \city{Kawasaki}
  \country{Japan}}
\email{hamada-naoki@jp.fujitsu.com}

\author{Keisuke Goto}
\affiliation{%
  \institution{Fujitsu Laboratories Ltd.}
  \streetaddress{4-1-1 Kamikodanaka, Nakahara-ku}
  \city{Kawasaki}
  \country{Japan}}
\email{goto.keisuke@jp.fujitsu.com}


\begin{abstract}
When and why can evolutionary multi-objective optimization (EMO) algorithms cover the entire Pareto set?
That is a major concern for EMO researchers and practitioners.
A recent theoretical study revealed that (roughly speaking) if the Pareto set forms a topological simplex (a curved line, a curved triangle, a curved tetrahedron, etc.), then decomposition-based EMO algorithms can cover the entire Pareto set.
Usually, we cannot know the true Pareto set and have to estimate its topology by using the population of EMO algorithms during or after the runtime.
This paper presents a data-driven approach to analyze the topology of the Pareto set.
We give a theory of how to recognize the topology of the Pareto set from data and implement an algorithm to judge whether the true Pareto set may form a topological simplex or not.
Numerical experiments show that the proposed method correctly recognizes the topology of high-dimensional Pareto sets within reasonable population size.
\end{abstract}

%
%

\ccsdesc[500]{Applied computing~Multi-criterion optimization and decision-making}
\ccsdesc[500]{Mathematics of computing~Evolutionary algorithms}
\ccsdesc[500]{Mathematics of computing~Nonconvex optimization}
\ccsdesc[500]{Mathematics of computing~Geometric topology}


\keywords{multi-objective optimization, continuous optimization, topological data analysis, persistent homology}

\maketitle

\section{Introduction}\label{sec:introduction}
Evolutionary multi-objective optimization (EMO) algorithms have celebrated successes ranging from engineering to science.
In recent years, encouraged by the growth of algorithms and computing environments, practitioners become to formalize their applications as \emph{many-objective} problems that have four or more objective functions~\cite{Chand15}.
Finding the many-objective Pareto set (the solution set in the decision space) is a challenging task since its dimensionality gets higher as the number of objectives increases.
Empirical studies reported that decomposition-based EMO algorithms such as MOEA/D~\cite{Zhang07,Wang16} and NSGA-III~\cite{Deb14} can provide a good covering of the entire Pareto set in the many-objective case while most of the other approaches worsen their performance.
Mathematical conditions ensuring when the decomposition-based approach works well are useful knowledge to both practitioners and researchers, but are still unclear.

Last year, one theoretical study~\cite{Hamada17b} found a problem class where the decomposition-based approach can easily cover the entire Pareto set, provided that the optima of scalarized objective functions are obtainable.
In such a problem, called a \emph{simple} problem, the Pareto set forms a topological simplex and a $k$-face of the simplex is the Pareto set of a subproblem optimizing $(k+1)$ objective functions of the original problem.
This topological structure ensures that if the weight for scalarizing objective functions is chosen from a face of the simplex of possible weights, then a solution on the corresponding face of the Pareto set is obtained.
Thus, a decomposition-based EMO algorithm with weights chosen to cover the simplex is guaranteed to cover the entire Pareto set of a simple problem (if the optima of the scalarized objective function for each weight are found).

Given an optimization problem, judging whether the problem is simple or not is an important task, but techniques to do it have not been established.
In this paper, we develop a simplicity test that respects the following nature of problems:
\begin{description}
 \item[Black-box] Recent EMO applications involve simulations in evaluating objective functions. Our method works in a purely data-driven manner and does not rely on the mathematical expression of objective functions.
 \item[Many-objective/variable] The Pareto set may be a 4--10D surface living in the 10--100D decision space. Our method employs persistent homology~\cite{Edelsbrunner02} to extract some topological features of such a surface.
\end{description}

The rest of this paper is organized as follows.
\Cref{sec:preliminaries} defines mathematical notions and their notations used through the paper.
\Cref{sec:proposal} proposes a data-driven method to test the simplicity of a given problem.
\Cref{sec:experiments} conducts numerical experiments to evaluate the proposed method.
\Cref{sec:discussion} gives discussion on the results.
\Cref{sec:conclusions} concludes the paper and discusses future work.

\section{Preliminaries}\label{sec:preliminaries}
\subsection{Multi-Objective Optimization}\label{sec:mop}
Throughout this paper, we consider the following optimization problem with $n$ variables and $m$ objective functions:
\begin{equation}\label{eq:MCOP}
 \minimize_{x \in X \paren{\subseteq \R^n}} f(x) := (f_1(x), \ldots, f_m(x)).
\end{equation}
We will also consider problems optimizing some of the objective functions.
To denote such problems, we treat a problem~\Cref{eq:MCOP} as a set of objective functions $f = \Set{f_1, \dots, f_m}$ and abuse set operations for describing relations among problems.
If two problems, say $f$ and $g$, satisfy $g \subproblemeq f$ in terms of the set inclusion, then we say $g$ is a \emph{subproblem} of $f$ and $f$ is a \emph{superproblem} of $g$.

Given a problem $f$, the \emph{Pareto set} $X^*(g)$ and the \emph{weak Pareto set} $X^{\mathrm{w}}(g)$ of a subproblem $g\subproblemeq f$ are defined by
\begin{align*}
X^*(g) &:= \Set{x\in X | \forall y\in X: \paren{g(x)=g(y) \lor \exists f_i \in g: f_i(x) < f_i(y)}},\\
X^{\mathrm{w}}(g) &:= \Set{x\in X | \forall y\in X,\exists f_i \in g: f_i(x) \le f_i(y)}.
\end{align*}
We call $fX^*(g):=\set{f(x)\in\R^{\card{f}} | x\in X^*(g)}$ the \emph{Pareto set image} of $g$.
In particular, $gX^*(g)$ is called the \emph{Pareto front} of $g$.

\subsection{Simple Problems}\label{sec:simple-problems}
\begin{figure*}[t!]
\centering%
\begin{tabular}{ccc}
\mpg[0.3]{
\begin{eqnarray*}
\minimize_{x_1,x_2 \in [-2,2]}\\
f_1(x) =& \hspace{-1.5em} x_1^2 &+\ 3(x_2 - 1)^2\\
f_2(x) =& 2(x_1 - 1)^2             &+\quad x_2^2\\
f_3(x) =& 3(x_1 + 1)^2             &+\ 2(x_2 + 1)^2
\end{eqnarray*}}
\ig[0.3]{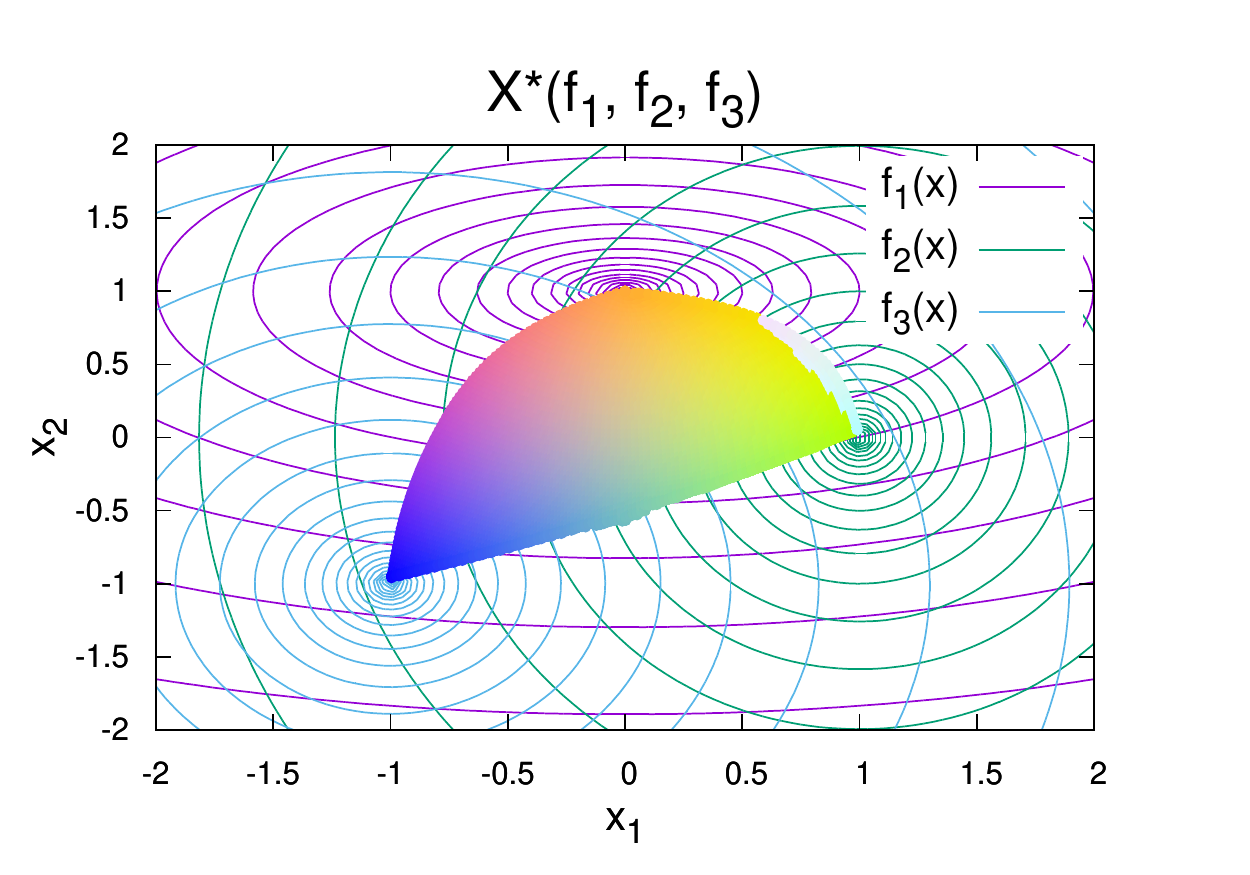}
\mpg[0.3]{\ }\\
\ig[0.3]{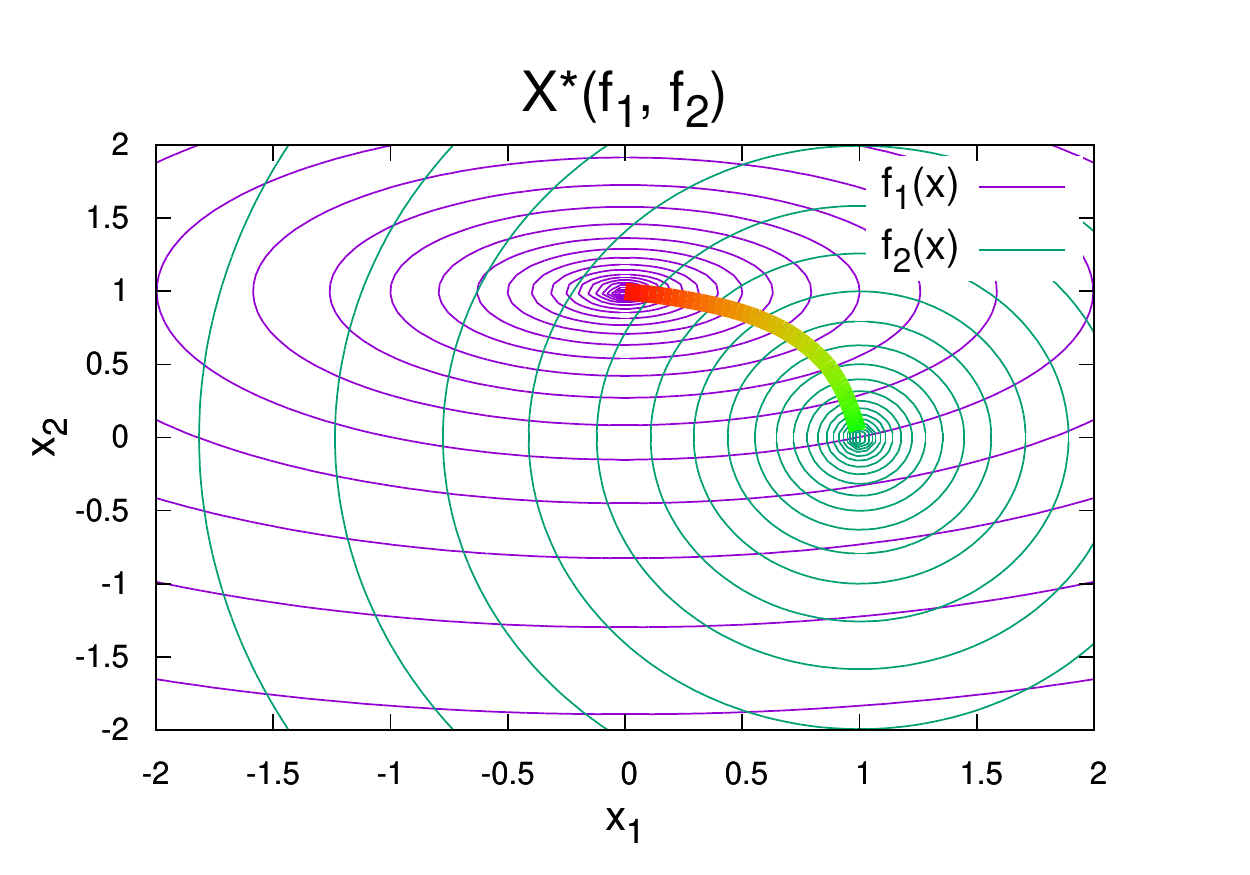}
\ig[0.3]{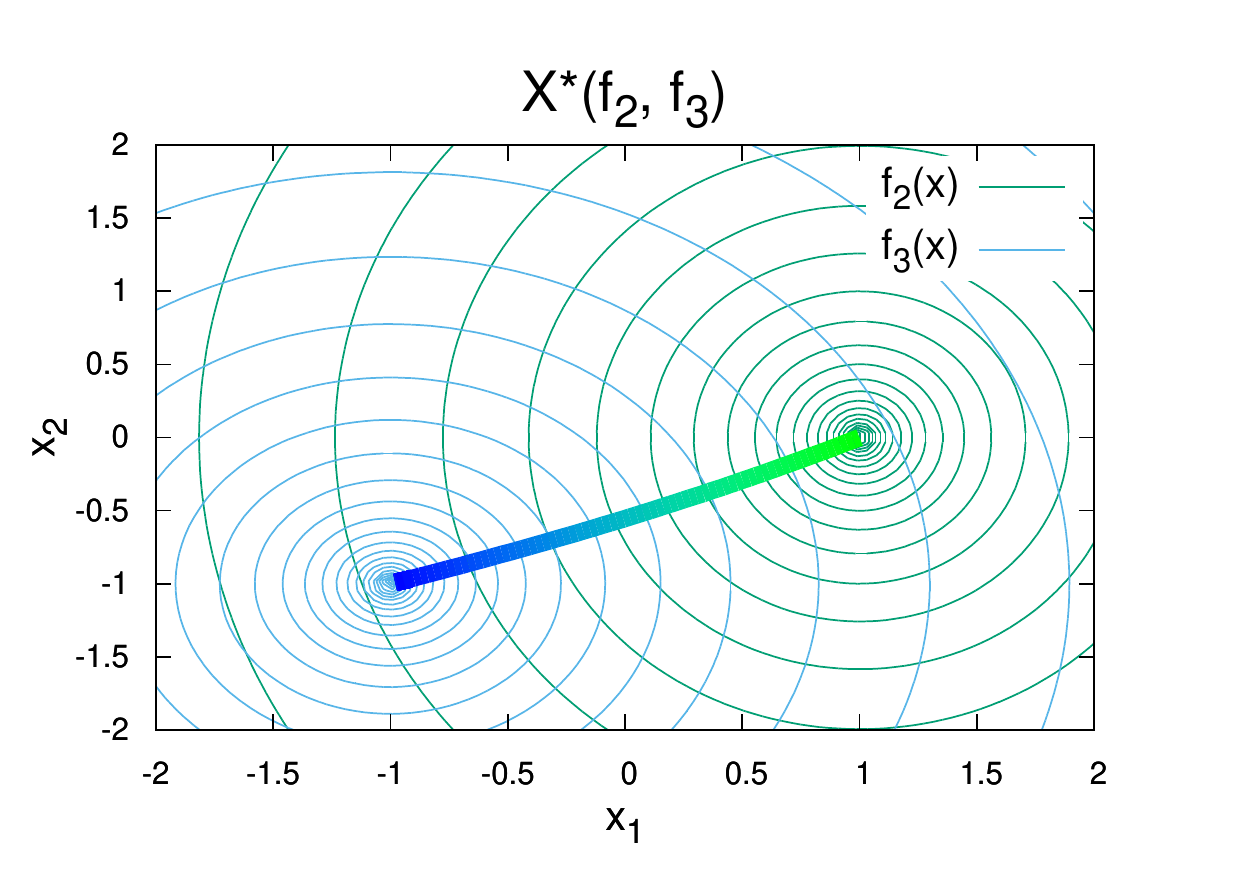}
\ig[0.3]{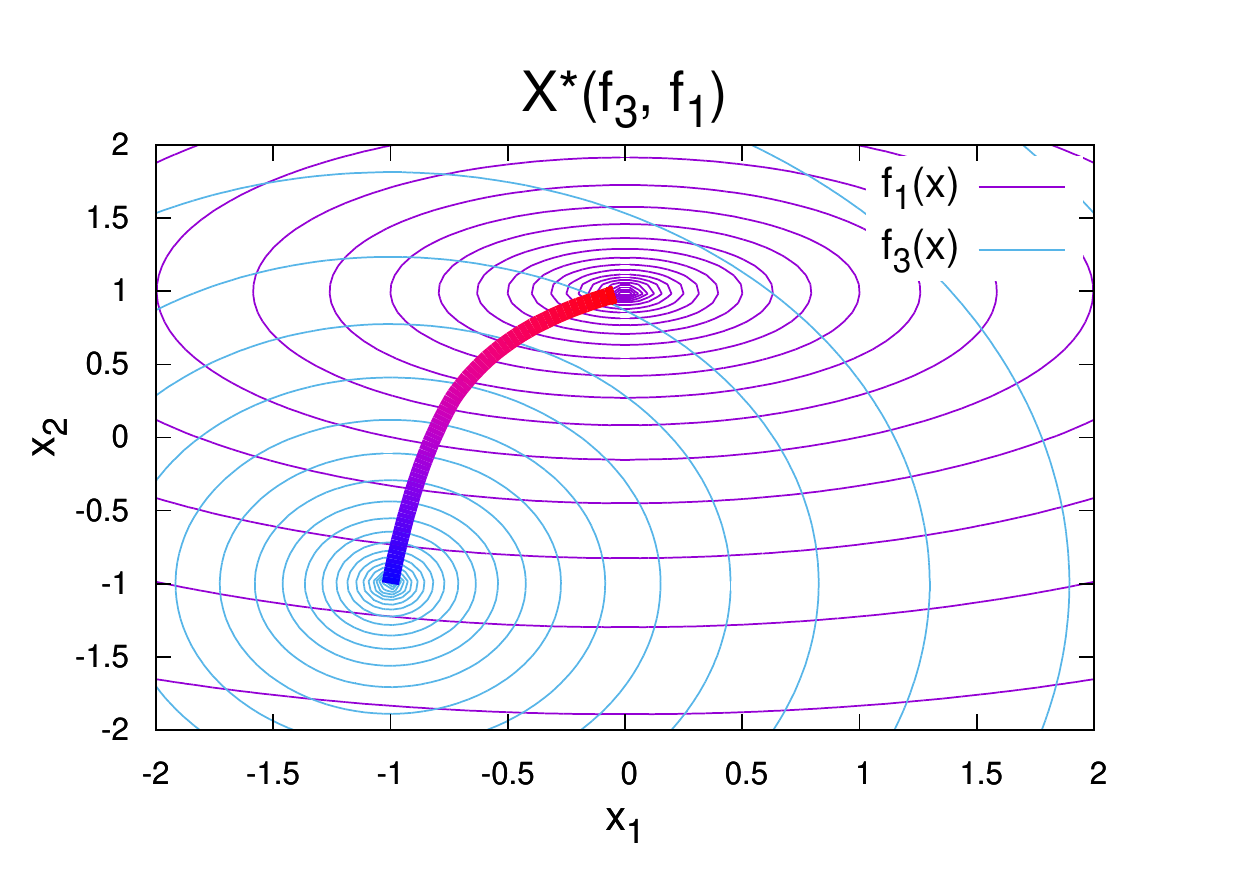}\\
\ig[0.3]{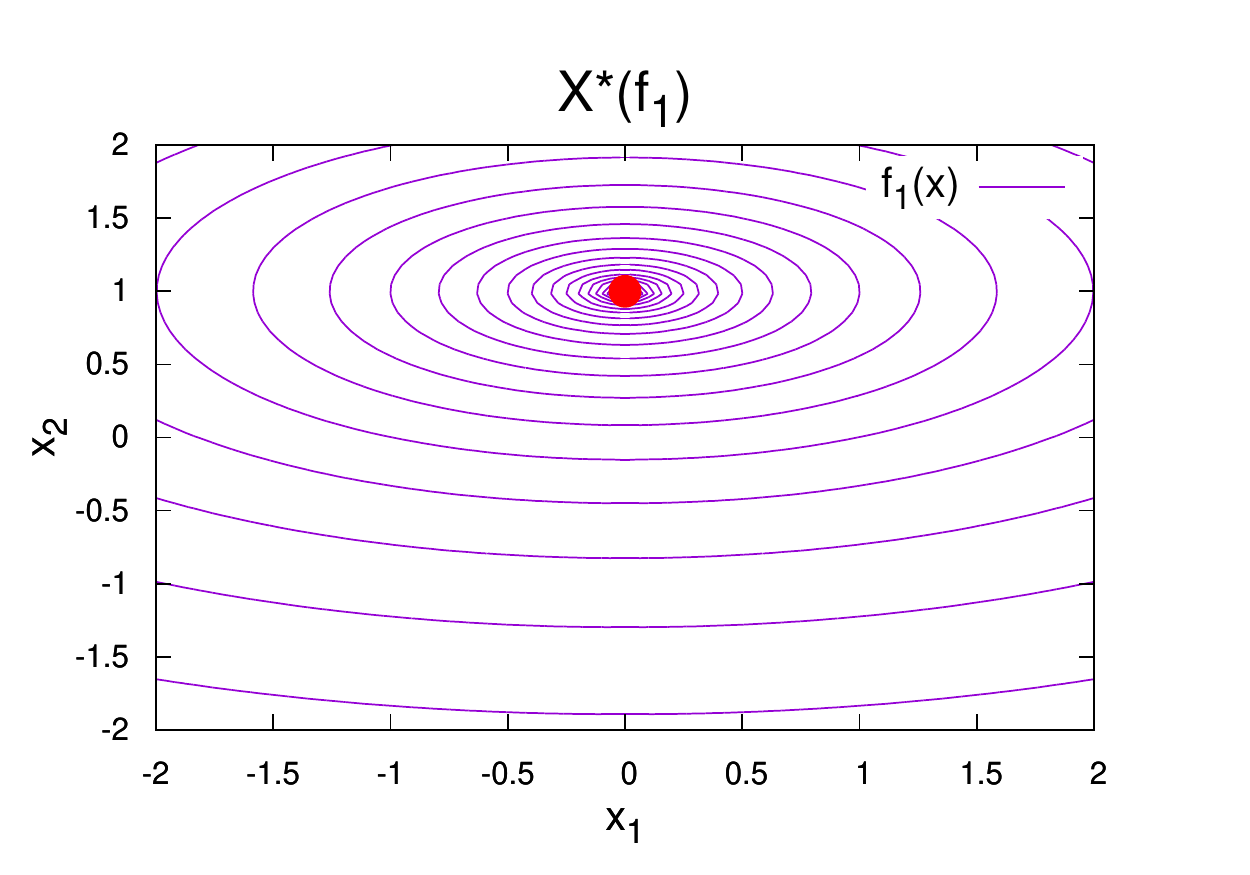}
\ig[0.3]{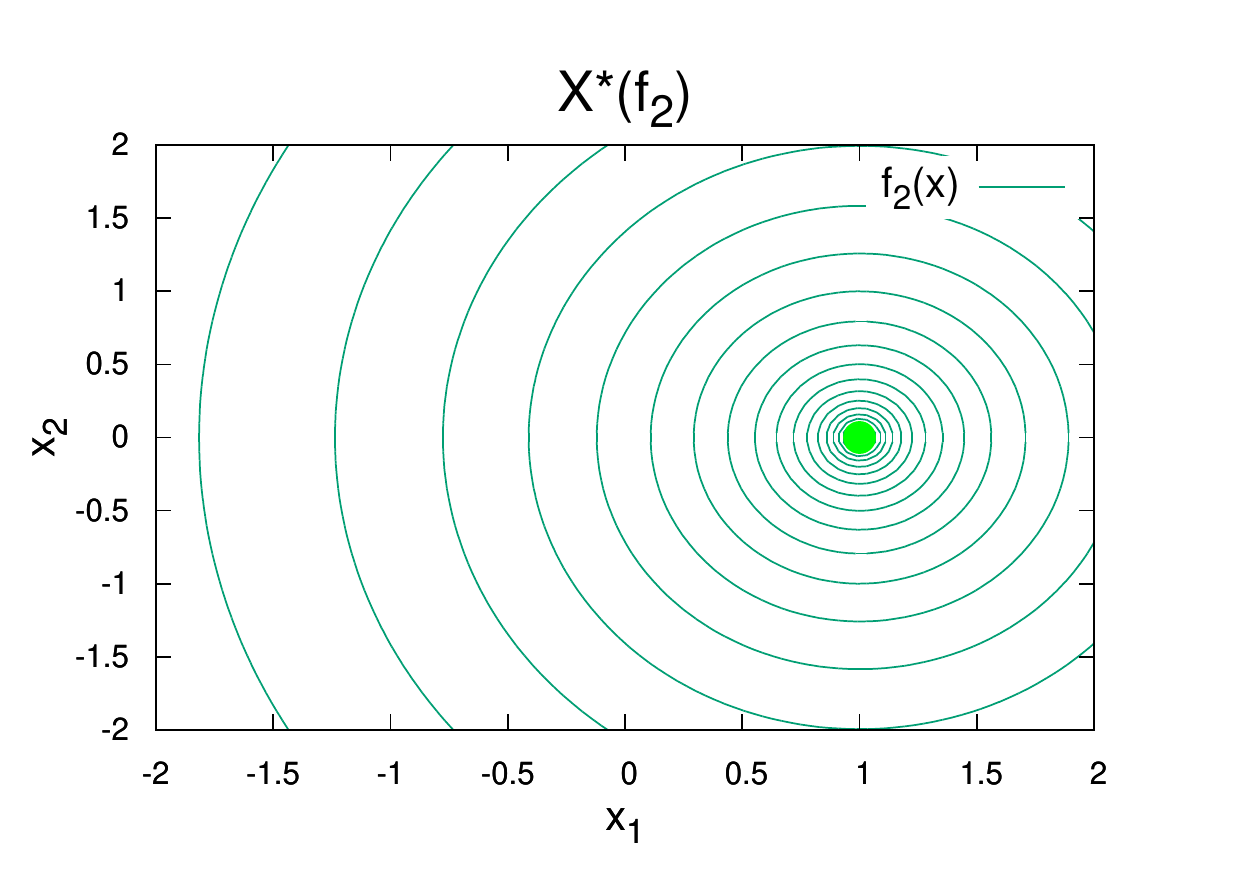}
\ig[0.3]{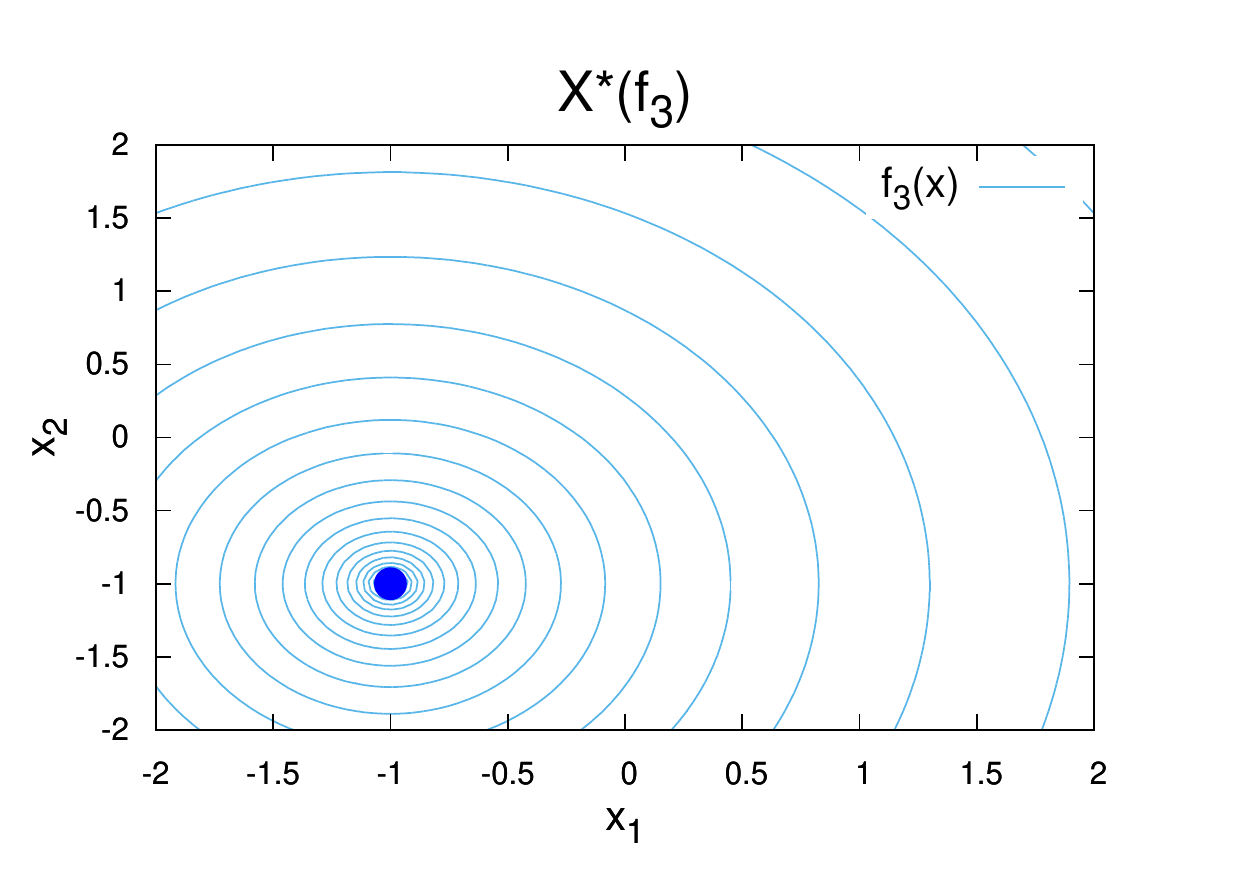}\\
\mpg[0.9]{\caption{The Pareto sets of subproblems of a simple problem. The simplicity condition (S1) states a 3-objective Pareto set is a curved 2-simplex (triangle), a 2-objective Pareto set is a curved 1-simplex (line) and a 1-objective Pareto set is a curved 0-simplex (point). The simplicity condition (S2) states the same for Pareto fronts.}
\label{fig:simple}}
\end{tabular}
\end{figure*}
We introduce the definition of the simple problem and its properties.
\begin{defn}[Simple problem~\cite{Hamada17b}]\label{def:simplicity}
A problem $f$ is \emph{simple} if each subproblem $g \subproblemeq f$ satisfies both the following conditions:
If $g$ has $k=\card{g}$ objectives, then
\begin{description}
 \item[(S1)] the Pareto set $X^*(g)$ of problem $g$ is homeomorphic to the standard $(k-1)$-simplex $\Delta^{k-1}$, i.e., $X^*(g) \homeo \Delta^{k-1}$;
 \item[(S2)] the objective mapping restricted to the Pareto set, $g|_{X^*(g)}: X^*(g) \to \R^k$, is topological embedding, i.e., $X^*(g) \homeo g X^*(g)$
\end{description}
where $\Delta^{k-1} := \set{x \in [0,1]^k | \sum_{i=1}^k x_i =1}$.
\end{defn}
By definition, every subproblem of a simple problem is again simple.
The simplicity conditions, (S1) and (S2), imply that solution sets of subproblems of a simple problem are well-formed in the sense described below.

\begin{figure*}[t!]
\centering%
\subfloat[Pareto sets $X^*(g)\ (g \subproblemeq f)$ ]{\includegraphics[width=0.475\hsize]{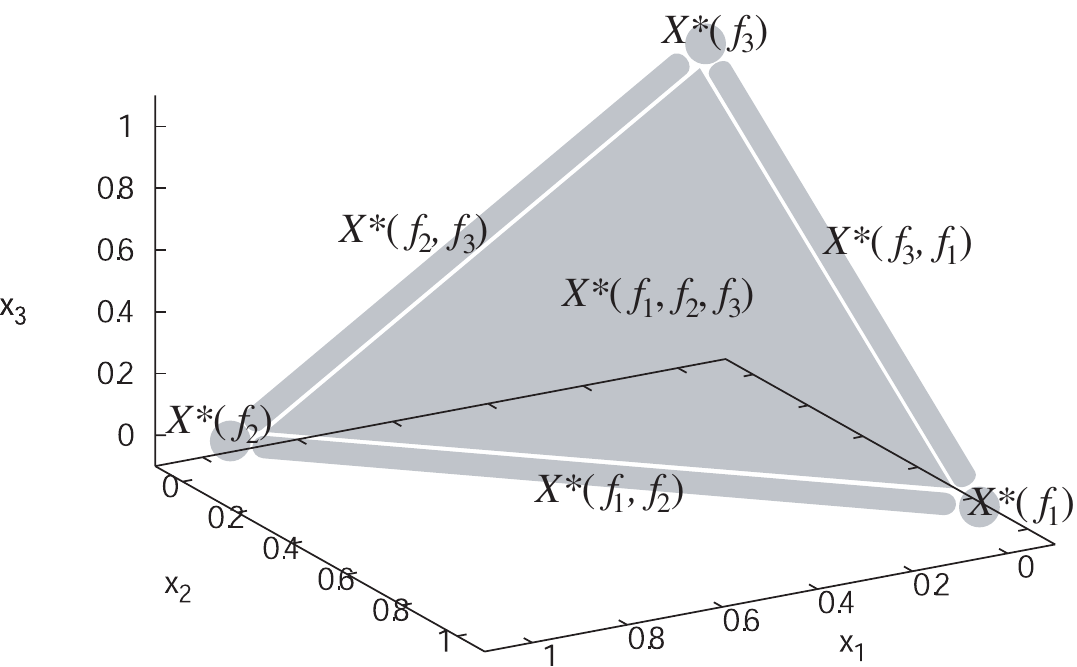}}\label{fig:simple-problem_X}
\subfloat[Pareto set images $f X^*(g)\ (g \subproblemeq f)$]{\includegraphics[width=0.475\hsize]{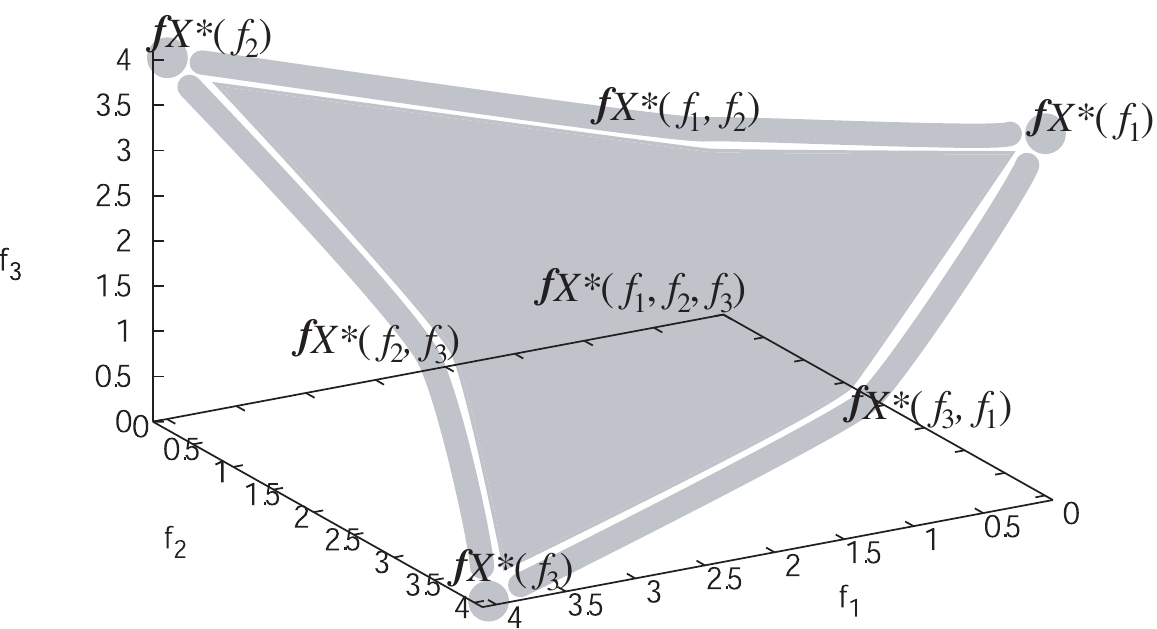}}\label{fig:simple-problem_F}
\caption{The face relation of Pareto sets  (their images) of a simple problem $f = \Set{f_1, f_2, f_3}$.}\label{fig:simple-problem}
\end{figure*}

First of all, the weak Pareto set coincides with the Pareto set.
\begin{thm}[Weak Pareto = Pareto~\cite{Hamada17b}]\label{thm:WP-is-P}
Let $f$ be a simple problem. Every subproblem $g \subproblemeq f$ satisfies 
\[
X^{\mathrm{w}}(g) = X^*(g).
\]
\end{thm}
Next, the Pareto set, the Pareto front and the Pareto set image are all homeomorphic to a simplex as shown in~\Cref{fig:simple}.
\begin{thm}[Topological type~\cite{Hamada17b}]\label{thm:topological-simplex}
Let $f$ be a simple problem.
Every subproblem $g \subproblemeq f$ satisfies
\begin{equation*}
X^*(g) \homeo g X^*(g) \homeo f X^*(g) \homeo \Delta^{\card{g}-1}.
\end{equation*}
\end{thm}
Finally, these simplices are shown to have the face relation as depicted in~\Cref{fig:simple-problem}.
\begin{thm}[Face relation~\cite{Hamada17b}]\label{thm:face-relation}
For a simple problem $f$ and any subproblem $g \subproblemeq f$, the following relationships hold:
\begin{align*}
\boundary X^*(g)		&= \bigsqcup_{h \subset g} \Int X^*(h),\\
\boundary f X^*(g)		&= \bigsqcup_{h \subset g} \Int f X^*(h),\\
f \boundary X^*(g)		&= \boundary f X^*(g),\\
f \Int X^*(g)			&= \Int f X^*(g)
\end{align*}
where $\sqcup$ is the disjoint union, and $\Int$ and $\boundary$ are the interior and the boundary of a topological manifold with boundary, respectively.
\end{thm}

\subsection{Scalarization and Decomposition}\label{sec:scalarization}
The topological structure of the Pareto sets and the Pareto fronts described in~\Cref{thm:topological-simplex,thm:face-relation} performs a crucial role when a decomposition-based EMO algorithm solves a problem.
It induces a natural stratification of the Pareto set and the Pareto front.
That is, the Pareto set $X^*(f)$ (resp.\ the Pareto front $fX^*(f)$) is decomposed into manifolds without boundary such that each stratum (i.e., a manifold without boundary) is the interior of the Pareto set $\Int X^*(g)$ (resp.\ its image $\Int fX^*(g)$) of a subproblem $g\subproblemeq f$.
Therefore, we can numerically compute the stratification by solving each subproblem.
Points spreading over all strata can be a good covering of the Pareto set and the Pareto front.

To see why this structure enables decomposition-based EMO algorithms to cover the Pareto set and the Pareto front, consider the weighted Chebyshev-norm scalarization defined by
\begin{equation}\label{eq:scalarization}
\minimize_{x \in X} f_w(x) := \max_i w_i \paren{f_i(x) - z_i}
\end{equation}
where the weight $w=(w_1, \dots, w_m)$ is chosen from $\Delta^{m-1}$ and the ideal point is fixed to be $z_i = \min_{x \in X} f_i(x)$.
Let $e_i$ be the $i$-th standard base in $\R^m$ whose $i$-th coordinate is one and the other coordinates are zero.
We denote the convex hull of points $p_1,\dots,p_k\in\R^l$ by
\[
[p_1,\dots,p_k] := \Set{p\in\R^l | p=\sum_{i=1}^k v_i p_i,\ v_i\in[0,1]}
\]
and rewrite the standard $(m-1)$-simplex as $\Delta^{m-1} = [e_1, \dots, e_m]$.
Using the notation of the weight-optima correspondence
\[
S(W) := \bigcup_{w \in W} X^*(f_w),
\]
a well-known fact of the optima to \Cref{eq:scalarization} can be written as
\begin{equation}\label{eq:Chebyshev-solution}
S([e_{i_1}, \dots, e_{i_k}]) = X^{\mathrm{w}}(f_{i_1}, \dots, f_{i_k}).
\end{equation}
for any choice of indices $i_1, \dots, i_k \in \Set{1, \dots, m}$ with an arbitrary number $1 \le k \le m$.
If the problem is simple, then we can go further: \Cref{thm:WP-is-P} extends \Cref{eq:Chebyshev-solution} to
\[
S([e_{i_1}, \dots, e_{i_k}]) = X^*(f_{i_1}, \dots, f_{i_k}),
\]
and \Cref{thm:face-relation} ensures
\[
S(\boundary [e_{i_1}, \dots, e_{i_k}]) = \boundary X^*(f_{i_1}, \dots, f_{i_k}).
\]
Therefore, a weight on each face gives a boundary point of each stratum with corresponding indices.

Unfortunately, the weighted Chebyshev-norm does NOT give the correspondence between the interiors:
\[
S(\Int [e_{i_1}, \dots, e_{i_k}]) \ne \Int X^*(f_{i_1}, \dots, f_{i_k}).
\]
This is also true for other existing scalarization methods including the weighted sum, the augmented Chebyshev-norm and PBI~\cite{Zhang07}.
Nevertheless, once boundary points of a stratum are obtained, we can find new weights corresponding to interior points of the stratum by interpolating the weights used for the boundary points.
Thus, if the weights are sampled over $[e_{i_1}, \dots, e_{i_k}]$ as most of the decomposition-based EMO algorithms do, then the optima of the corresponding scalarized objective functions practically often hit interior points of $X^*(f_{i_1}, \dots, f_{i_k})$.
This working principle applies to MOEA/D~\cite{Zhang07,Wang16}, NSGA-III~\cite{Deb14} and AWA~\cite{Hamada10,Shioda14,Shioda15a,Shioda15b}.
That is, those decomposition-based EMO algorithms are guaranteed to cover the Pareto set and the Pareto front of a simple problem if the set of global optima $X^*(f_w)$ of problem \Cref{eq:scalarization} can be found for each weight $w\in \Delta^{m-1}$.

\section{Simplicity Test}\label{sec:proposal}
The black-box optimization is a problem where the mathematical expressions of objective functions are not available.
The only thing we can do is to query the value of objective functions at a specified point.
In this case, the simplicity of the problem cannot be determined from the expressions, and instead we first run an EMO algorithm to obtain a finite sample of the Pareto set and then estimate the simplicity from the sample.
Unless we put strong assumptions, it is very difficult to derive affirmative results (the problem is surely simple!) from a finite sample.
We check whether the necessary conditions for (S1) and (S2) hold or not.
In other words, we seek evidence against the simplicity of a problem and if it is not found, then we regard the problem might be simple.

\subsection{Non-Simplicity Detection}\label{sec:theory}
We check the simplicity conditions (S1) and (S2) separately.
\subsubsection{Test for (S1) Violation}\label{sec:S1}
We would like to test if (S1) holds or not.
To seek a contradiction of $X^*(g) \homeo \Delta^{\card{g}-1}$ for some subproblem $g \subproblemeq f$, we examine the isomorphism of their homology groups:
\[
\mbox{Are there any $g$ and $i$ such that } H_i(X^*(g)) \not \cong H_i(\Delta^{\card{g}-1})?
\]

The left hand side of the above isomorphism depends on unknown information $X^*(g)$.
So we need to replace it with information constructed from available data
\[
X(g) := \mbox{a finite sample around the Pareto set $X^*(g)$}.
\]
We use, as $X(g)$, a set of non-dominated points found by an EMO algorithm and, as a substitute for $X^*(g)$, a $d$-Rips complex with simplex diameter $\delta$ which is a simplicial complex defined by
\[
K_\delta^d(g) := \Set{\sigma \subseteq X(g) | \card{\sigma} \le d+1,\ \forall p,q \in \sigma: \norm{p-q} \le \delta}.
\]
Hereafter, we simply denote it by $K(g)$.

We compute the homology group $H_i(K(g))$ of a complex $K(g)$ and use the following theorem to detect the violation of the simplicity condition (S1).
\begin{thm}[Test for (S1) violation]\label{thm:S1}
Let $f$ be a problem and for all $g\subproblemeq f$, $K(g)$ be a simplicial complex and $\gr{K(g)}$ be its geometric realization such that $\gr{K(g)}$ is homotopy equivalent to $X^*(g)$, i.e., $\gr{K(g)} \simeq X^*(g)$.
If there exist some $g$ and $i$ that satisfy one of the following conditions, then the problem $f$ does not satisfy (S1).
\begin{itemize}
\item $H_i(K(g)) \not \cong \Z$ for $i=0$,
\item $H_i(K(g)) \not \cong 0$ for $i \ne 0$
\end{itemize}
where $H_i$ denotes the $i$-th simplicial homology group with $\Z$-coefficient and $\cong$ denotes the group isomorphism.
\end{thm}
\begin{prf}
By the topological invariance of homology, we have
\begin{equation}\label{eq:homology}
\begin{split}
X^*(g) \homeo \Delta^{\card{g}-1} \Rightarrow H_i(X^*(g)) &\cong H_i(\Delta^{\card{g}-1})\\
&\cong
\begin{cases}
 \Z &(i=0),\\
 0  & (i \ne 0).
\end{cases}
\end{split}
\end{equation}
By the homotopy invariance of homology, we have
\begin{equation}\label{eq:homotopy}
\gr{K(g)} \simeq X^*(g) \Rightarrow H_i(K(g)) \cong H_i(X^*(g)).
\end{equation}
Combining \Cref{eq:homology} and \Cref{eq:homotopy}, we have the theorem.
\end{prf}
Note that this is just a condition for denying (S1); checking its converse does not ensure that (S1) holds.

\subsubsection{Test for (S2) Violation}\label{sec:S2}
To contradict the simplicity condition (S2), we need to find that the restriction $g|_{X^*(g)}: X^*(g) \to \R^{\card{g}}$ is not a topological embedding.
For computation, we triangulate its domain $X^*(g)$ and range $g X^*(g)$.
We choose triangulations, i.e. homeomorphisms $\psi: \gr{K} \to X^*(g)$ and $\varphi: g X^*(g) \to \gr{L}$, such that there exists a simplicial approximation $\hg: K \to L$ to a continuous map $\varphi g \psi: \gr{K} \to \gr{L}$.
Then, we check whether $\hg$ is isomorphism or not in terms of a simplicial map.
\begin{thm}[Test for (S2) violation]\label{thm:S2}
Let $f$ be a problem and for all $g\subproblemeq f$, $K$ be a triangulation of $X^*(g)$ such that $K$ is a subdivision of $K(g)$.
Let $\sigma=[x_1, \dots, x_k], \tau=[y_1, \dots, y_l]$ such that $\sigma,\tau \in K(g)$, $\sigma \ne \tau$.
If the following system of linear equations has a solution, then problem $f$ does not satisfy the simplicity condition (S2):
\begin{equation}\label{eq:S2}
\left \{
\begin{split}
& \sum_{i=1}^k a_i g(x_i) = \sum_{j=1}^l b_j g(y_j),\\
& \sum_{i=1}^k a_i = \sum_{j=1}^l b_j = 1,\\
& a_i, b_j > 0.
\end{split}
\right.
\end{equation}
\end{thm}
\begin{prf}
Since a simplicial approximation is homotopic to the original mapping, it holds that
\begin{equation}\label{eq:simplicial-approximation}
\begin{split}
g|_{X^*(g)} \text{ topological embedding} &\Leftrightarrow \varphi g \psi \text{ homeomorphism}\\
 &\Rightarrow \hg \text{ isomorphism}\\
 &\Leftrightarrow \hg \text{ injective}.
\end{split}
\end{equation}
Thus, checking the non-injectivity of $\hg$ is enough to detect the (S2) violation.

The right hand side of \Cref{eq:simplicial-approximation} depends on an unknown mapping $\hg: K \to L$.
We need to approximate it with a mapping constructed from a finite sample $X(g)$.
Using the complex $K(g)$ in the decision space $\R^n$ constructed above, let us consider in the objective space $\R^m$ the set of convex hulls of vertex sets of its simplices
\[
L(g) = \Set{ \underbrace{[g(x_1), \dots, g(x_k)]}_{\text{a convex hull in }\R^m} | \underbrace{[x_1, \dots, x_k]}_{\text{a simplex in }\R^n} \in K(g)}.
\]
This does not necessarily become a simplicial complex, but if it does, the mapping
\begin{equation}\label{eq:simplicial-map}
\hg': \left \{
\begin{array}{ccc}
 K(g) & \to & L(g),\\
 {[x_1, \dots, x_k]} & \mapsto & [g(x_1), \dots, g(x_k)]
\end{array}
\right.
\end{equation}
is a simplicial map and it holds that
\[
\paren{K \text{ subdivision of } K(g)\ \land\ \hg \text{ injective}} \Rightarrow \hg' \text{ injective}.
\]
$L(g)$ is a simplicial complex and $\hg'$ is injective if and only if
\[
\Int \hg'(\sigma) \cap \Int \hg'(\tau) = \emptyset \quad\mbox{for all } \sigma, \tau \in K(g)\ (\sigma \ne \tau).
\]
Combining this condition with \Cref{eq:simplicial-approximation}, we have the theorem.
\end{prf}

\subsection{Diameter Determination}\label{sec:algorithm}
\begin{figure*}[t!]
\centering%
\subfloat[Scatter plot]{\includegraphics[width=0.495\hsize]{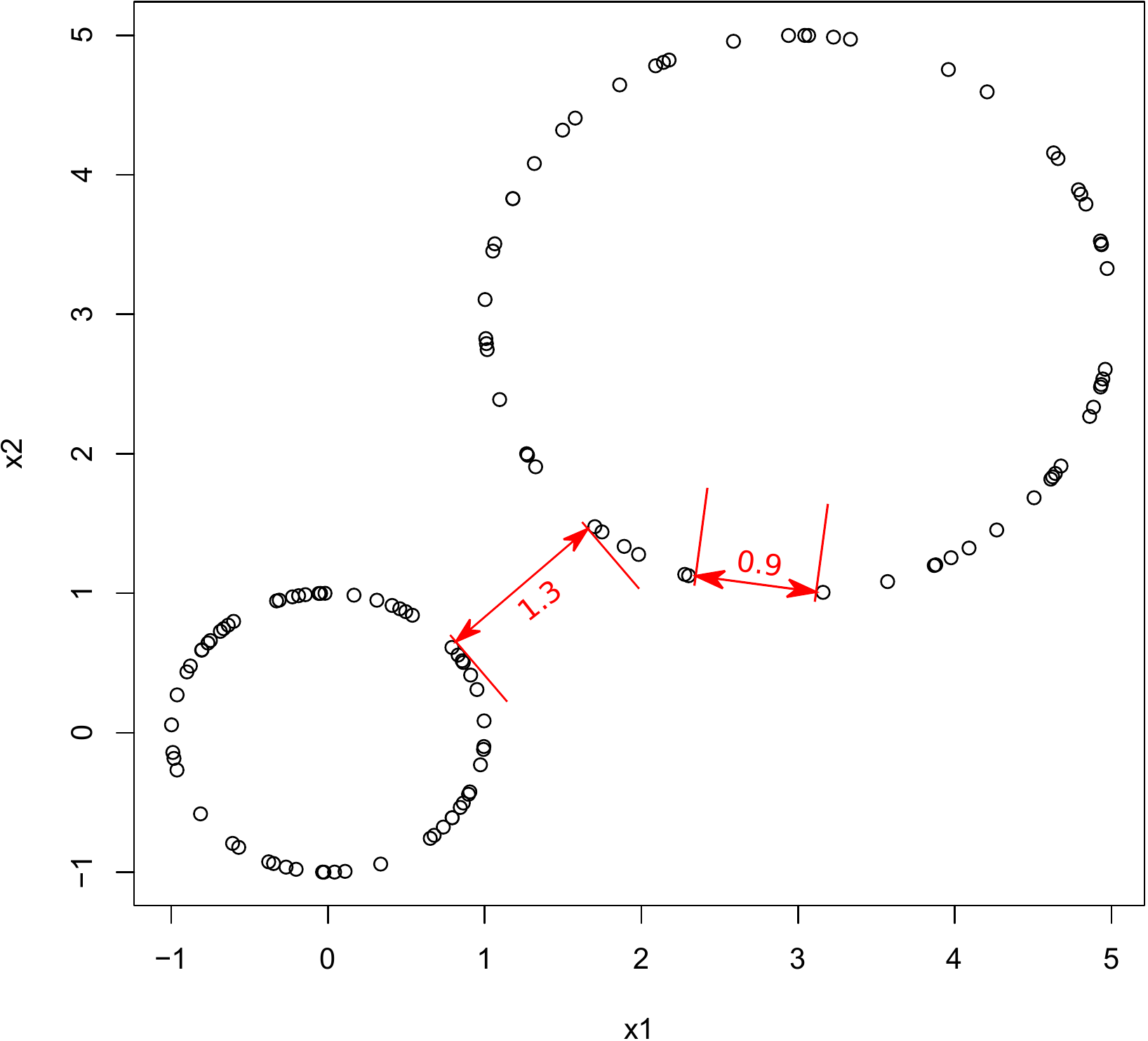}\label{fig:Circles}}
\subfloat[Persistent diagram ($\bullet$: 0-cycle, \textcolor{red}{$\triangle$}: 1-cycle)]{\includegraphics[width=0.495\hsize]{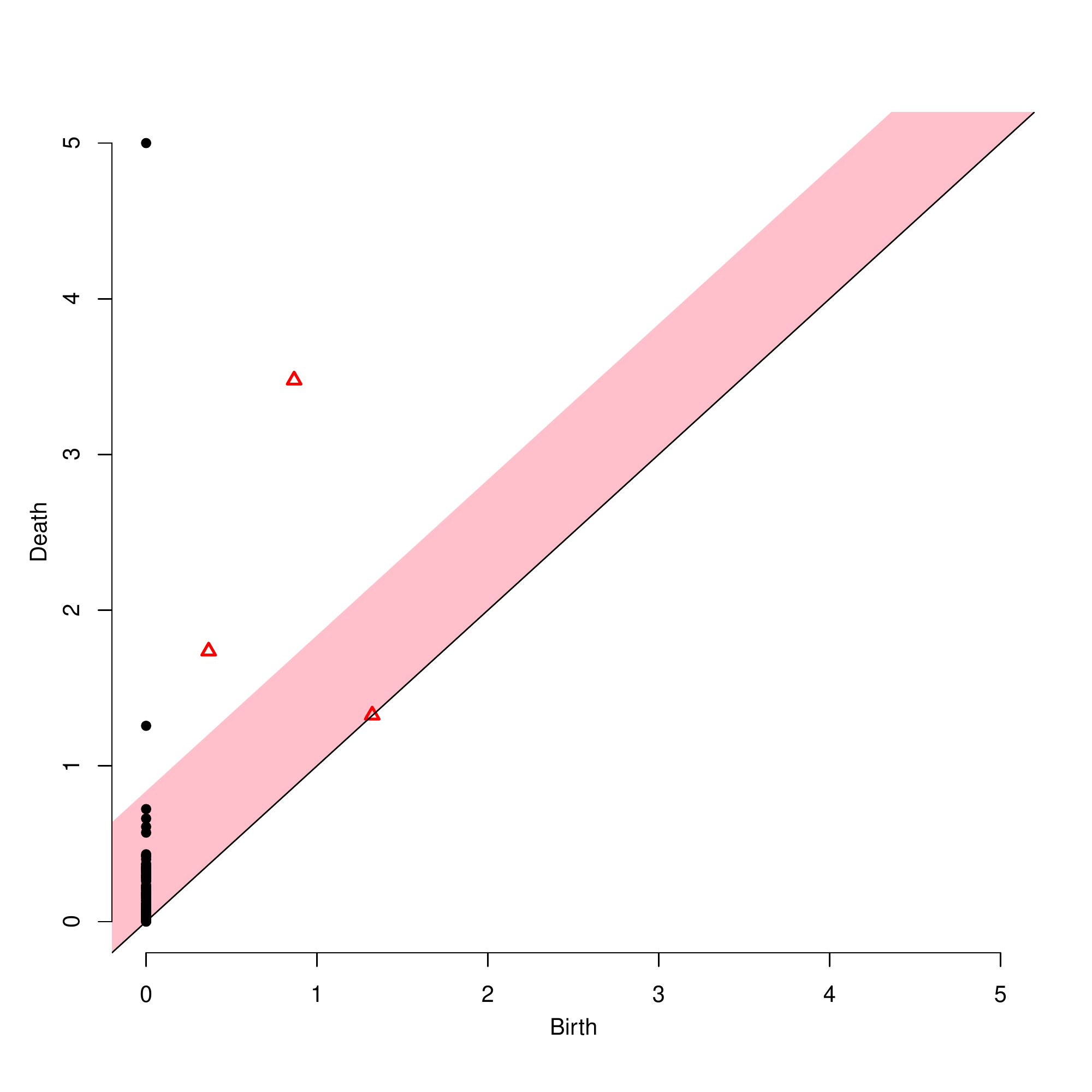}\label{fig:Circles-PD}}
\caption{Diameter determination via a scatter plot and a persistent diagram.}
\end{figure*}
Usually, we do not know the true Pareto set $X^*(g)$ and a triangulation $K$ of it.
It is not easy to construct a Rips complex $K(g)$ such that it satisfies the assumptions
\begin{itemize}
 \item $\gr{K(g)}$ is homotopy equivalent to $X^*(g)$,
 \item $K$ is a subdivision of $K(g)$.
\end{itemize}
Here we develop a method to construct such a complex.

\subsubsection{Via a Scatter Plot}
Let us first intuitively grasp the idea by using a synthetic dataset shown in~\Cref{fig:Circles}.
The underlying topology behind this sample is $S^1 \sqcup S^1$.
When one constructs a Rips complex from a sample and tries to find an appropriate diameter $\delta$ to recover the true topology.
Now, the farthest adjacent points in the large circle in~\Cref{fig:Circles} have distance 0.9.
Thus, we need the Rips diameter at least 0.9 to recover the large circle.
On the other hand, the distance between the small circle and the large circle is 1.3.
Thus, we need the Rips diameter less than 1.3 to prevent from connecting the small and large circles.
Consequently, the diameter $\delta$ such that the Rips complex becomes homeomorphic to $S^1 \sqcup S^1$ is
\begin{equation}\label{eq:diameter}
0.9 \le \delta < 1.3.
\end{equation}
This is intuitive but a 2D specific way since we cannot read the distance from scatter plots in general dimensions.

\subsubsection{Via a Persistent Diagram}
To determine an appropriate diameter in arbitrary dimensions, we employ persistent homology~\cite{Edelsbrunner02} that traces the topological changes of a growing complex.
Instead of using a fixed diameter of a Rips complex, this technique sweeps the diameter to get a \emph{filtration}
\[
K_{\delta=0} \subseteq \dots \subseteq K_{\delta} \subseteq \dots \subseteq K_{\delta=\infty}
\]
and tracks the birth and death of cycles in the filtration.
When a new hole surrounded by $k$-simplices arises in $K_\delta$, a generator of the $k$-homology group ($k$-cycle) is added, and when the hole is buried with a $(k+1)$-simplex, the cycle is identified with a unit and disappears.

The diameter $\delta = b_i$ at which a new cycle $i$ is added to a homology group is called the \emph{birth time} of $i$, and $\delta = d_i$ at which $i$ is disappeared is called the \emph{death time} of $i$.
The set of all birth-death pairs, $\Set{(b_i, d_i)}$, plotted on a 2D plane like \Cref{fig:Circles-PD} is called the \emph{persistent diagram}.
The red band along the diagonal indicates the 95\% confidence interval in which cycles are considered to be noise~\cite{Fasy14}.

There are many cycles but most of them are under the band and thus estimated to be sampling noise.
Only two 0-cycles and two 1-cycles exist above the band.
Their birth-death pairs are:
\begin{itemize}
\item 0-cycle $(0.0, 1.3)$ corresponding to the small circle,
\item 0-cycle $(0.0, 5.0)$ corresponding to the large circle\footnote{The death time 5.0 is due to the computation limit of the Rips diameter; the actual death time is $\infty$.},
\item 1-cycle $(0.4, 1.8)$ corresponding to the small circle,
\item 1-cycle $(0.9, 3.5)$ corresponding to the large circle.
\end{itemize}
Taking $\max b_i$ and $\min d_i$, all signal cycles alive between
\[
0.9 \le \delta < 1.3.
\]
Therefore, \Cref{fig:Circles-PD} tells the same condition \Cref{eq:diameter} as \Cref{fig:Circles}.
Since the persistent diagram is always 2D, we can use this method in arbitrary dimensions.
The proposed method uses the middle of lifetime $(\max b_i + \min d_i)/2$ as an estimate of the appropriate diameter.

\section{Numerical experiments}\label{sec:experiments}
This section demonstrates the effectiveness of the proposed method.
For several benchmark problems, we compute their solutions and construct their persistent diagrams.

\subsection{Settings}\label{sec:settings}
We used four benchmark problems described below:

\paragraph{(40, 6)-MED~\cite{Hamada11b}}
This 40-variable, 6-objective problem is mathematically proven to be a simple problem~\cite{Hamada17b}.
We used it to check that our method does not raise false positives of non-simplicity.
The Pareto set is known as
\begin{equation}\label{eq:MED}
X^*(f)=[e_1,\dots,e_6] \subset \R^{40},
\end{equation}
i.e., a 5-simplex spanned by the first six of the standard basis of $\R^{40}$, whose vertices are the optima of the individual objective functions.

\paragraph{(40, 6)-Gapped MED}
To test the detection of discontinuity of objective functions, which violates (S2), we introduced ``gaps'' that break the continuity of MED's objective functions as follows:
\[
\begin{split}
\minimize_{x \in \R^{40}}  f(x) &:= (f_1(x), \ldots, f_6(x)),\\
\text{where }	f_i(x) &= \begin{cases}
\frac{2}{3} g_i(x)               & \mbox{if } g_i(x) \le \frac{1}{2},\\
\frac{2}{3} g_i(x) + \frac{1}{3} & \mbox{if } g_i(x)  >  \frac{1}{2},\\
\end{cases}\\
		g_i(x) &= \paren{ \frac{1}{\sqrt{2}} \norm{ x - e_i } }^{p_i},\\
		e_i &= (\underbrace{0,\ldots,0,}_{i-1} 1 \underbrace{, 0,\ldots,0}_{40-i}),\\
		p_i &= \exp \paren{\frac{2i-7}{5}}.
\end{split}
\]
The Pareto set is the same as \Cref{eq:MED} but the objective functions are discontinuous on it.

\paragraph{(12, 3)-DTLZ5~\cite{Deb05b}}
This 12-variable, 3-objective problem tests the detection of another cause for violating (S2): objective functions are many-to-one on the Pareto set.
The Pareto set is analytically unknown but we can numerically compute it since the objective functions are unimodal.

\paragraph{(22, 3)-DTLZ7~\cite{Deb05b}}
The last problem with 22 variables and 3 objectives has a disconnected Pareto set that violates (S1).
The Pareto set is analytically unknown but we can numerically find it since the objective functions are unimodal.

We generated Pareto set samples in the following manner:
For MED and Gapped MED, 300 random points were drawn from the uniform distribution on their Pareto sets.
For DTLZ5 and DTLZ7, MOEA/D with population size 300 was used to approximate their Pareto sets.
The implementation of MOEA/D was employed from \texttt{jMetal~5.2}~\cite{Nebro15} with its default settings.

For each problem, we computed a persistent diagram and its 95\% confidence set.
All experiments are conducted 10 times on Xeon 3.5 GHz, 64 bit, 32 GB RAM using \texttt{R x64 3.4.3}~\cite{R15} and package \texttt{TDA 1.6}~\cite{TDA15}.
Since the size of a $d$-Rips complex grows exponentially with respect to its dimensionality $d$, we only computed 2-Rips complexes and up to 2-homologies.
We also restricted the computation of our linear equations \Cref{eq:S2} only for 5-simplices due to computational complexity.

\subsection{Results}\label{sec:results}
\begin{table*}
\caption{Results of simplicity test.}\label{tbl:results}
\begin{tabular}{lllrrr}
\toprule%
\bf Problem        & \multicolumn{2}{l}{\bf Ground Truth}                                   & \multicolumn{3}{l}{\bf Estimated}\\
                   & (S1)                                & (S2)                             & Average $\delta$ & (S1) unsatisfied & (S2) unsatisfied\\
\midrule%
(40,6)-MED         & \checkmark                           & \checkmark                       & 0.500            &  0               &  0\\
(40,6)-Gapped MED  & \checkmark                           & $\times$ ($f$ is discontinuous) & 0.500            &  0               &  0\\
(12,3)-DTLZ5       & \checkmark                           & $\times$ ($f$ is many-to-one)   & 0.438            &  2               & 10\\
(22,3)-DTLZ7       & $\times$ ($X^*(f)$ is disconnected) & \checkmark                       & 0.191            & 10               &  1\\
\bottomrule%
\end{tabular}
\end{table*}

The results are shown in \Cref{tbl:results}.
Most of the cases were correctly estimated.
There are few false positives: $2/10$ trials in the (S1) violation for DTLZ5 and $1/10$ trials in the (S2) violation for DTLZ7.
They came from sampling errors.
There is also a serious false negative: $10/10$ trials missed the (S2) violation for Gapped MED.
It implies our method does not have an ability to detect the discontinuity.

\section{Discussion}\label{sec:discussion}
\subsection{Detection Accuracy}
\begin{figure*}[t!]
\centering%
\subfloat[(40,6)-MED]{\includegraphics[width=0.25\hsize]{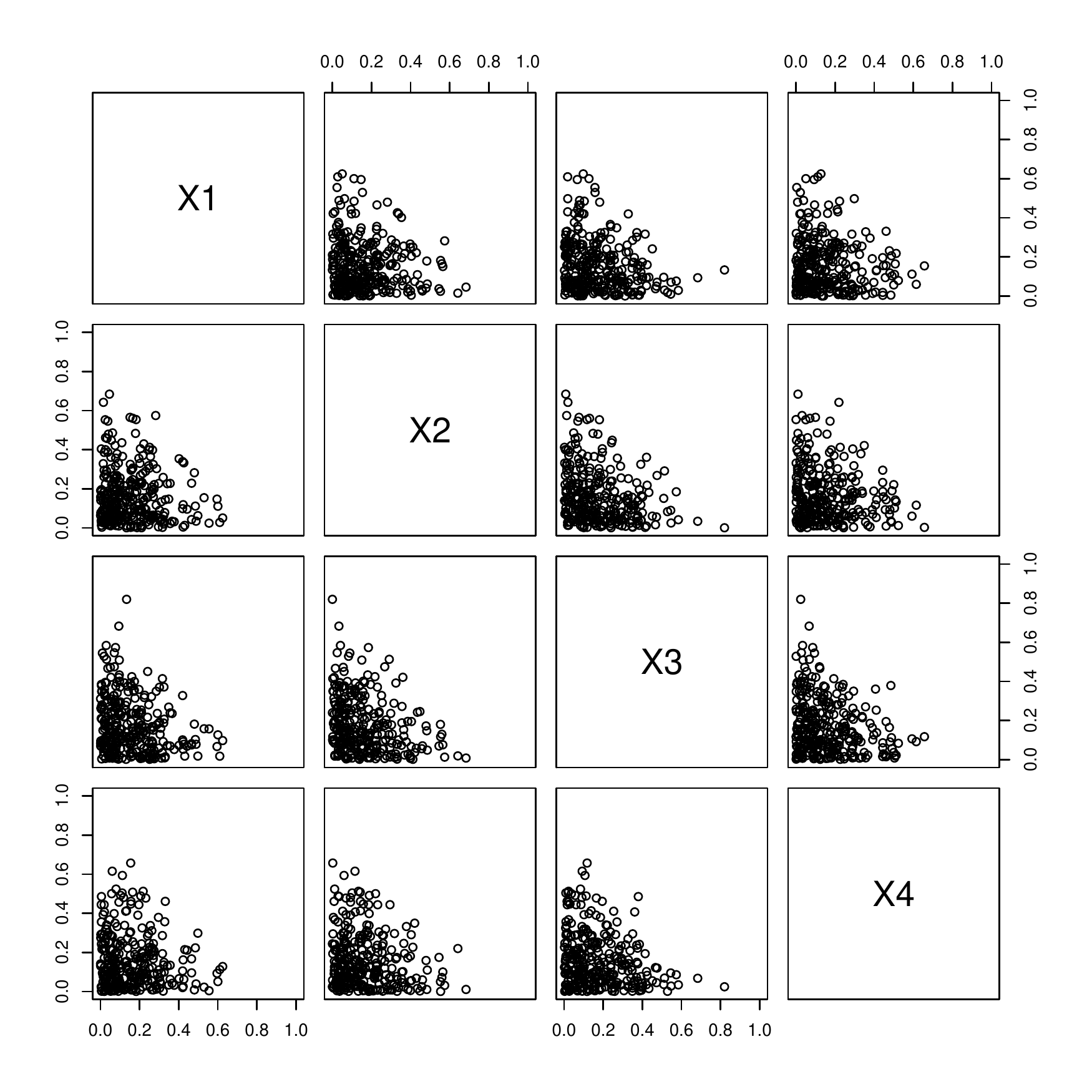}\includegraphics[width=0.25\hsize]{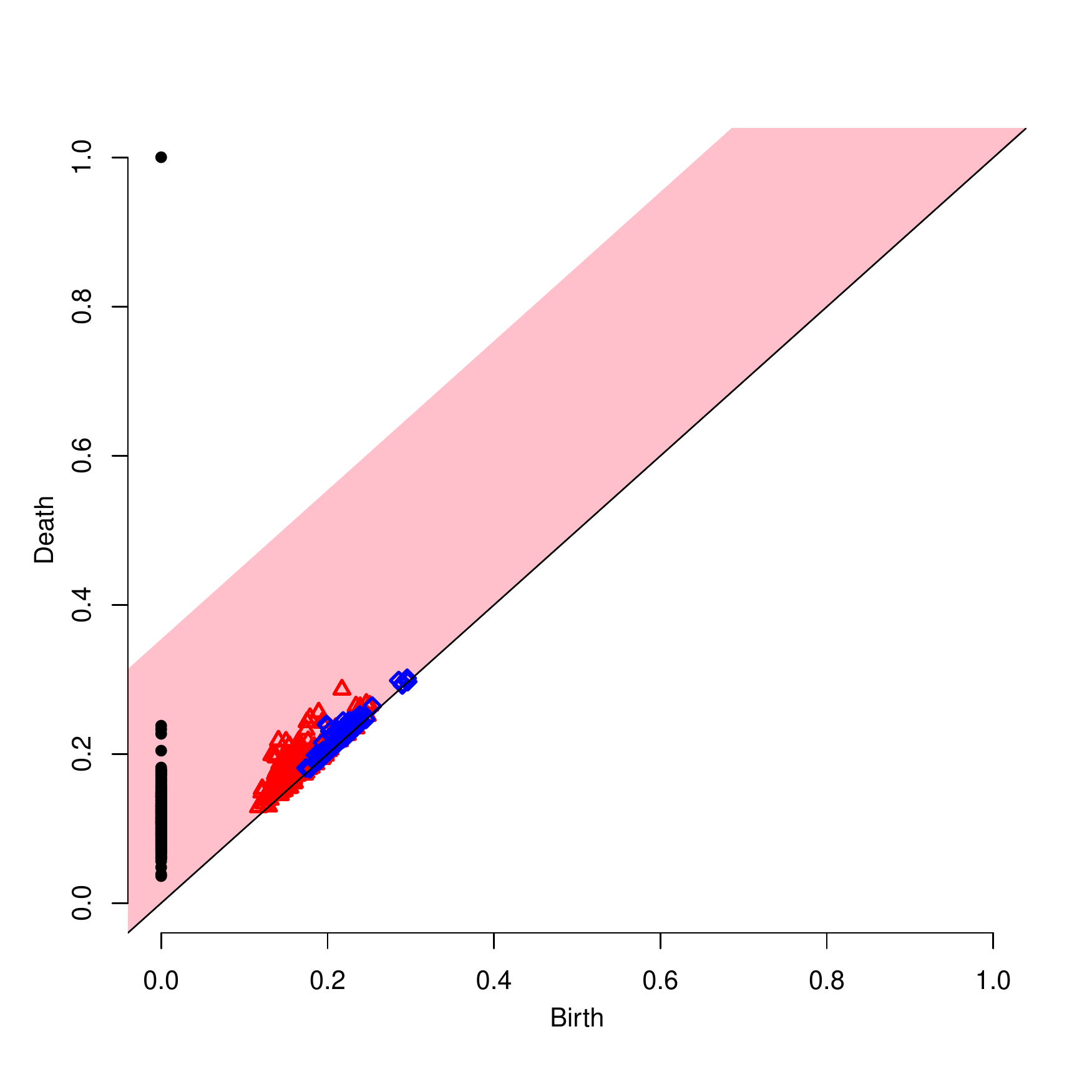}\label{fig:MED}}
\subfloat[(40,6)-Gapped MED]{\includegraphics[width=0.25\hsize]{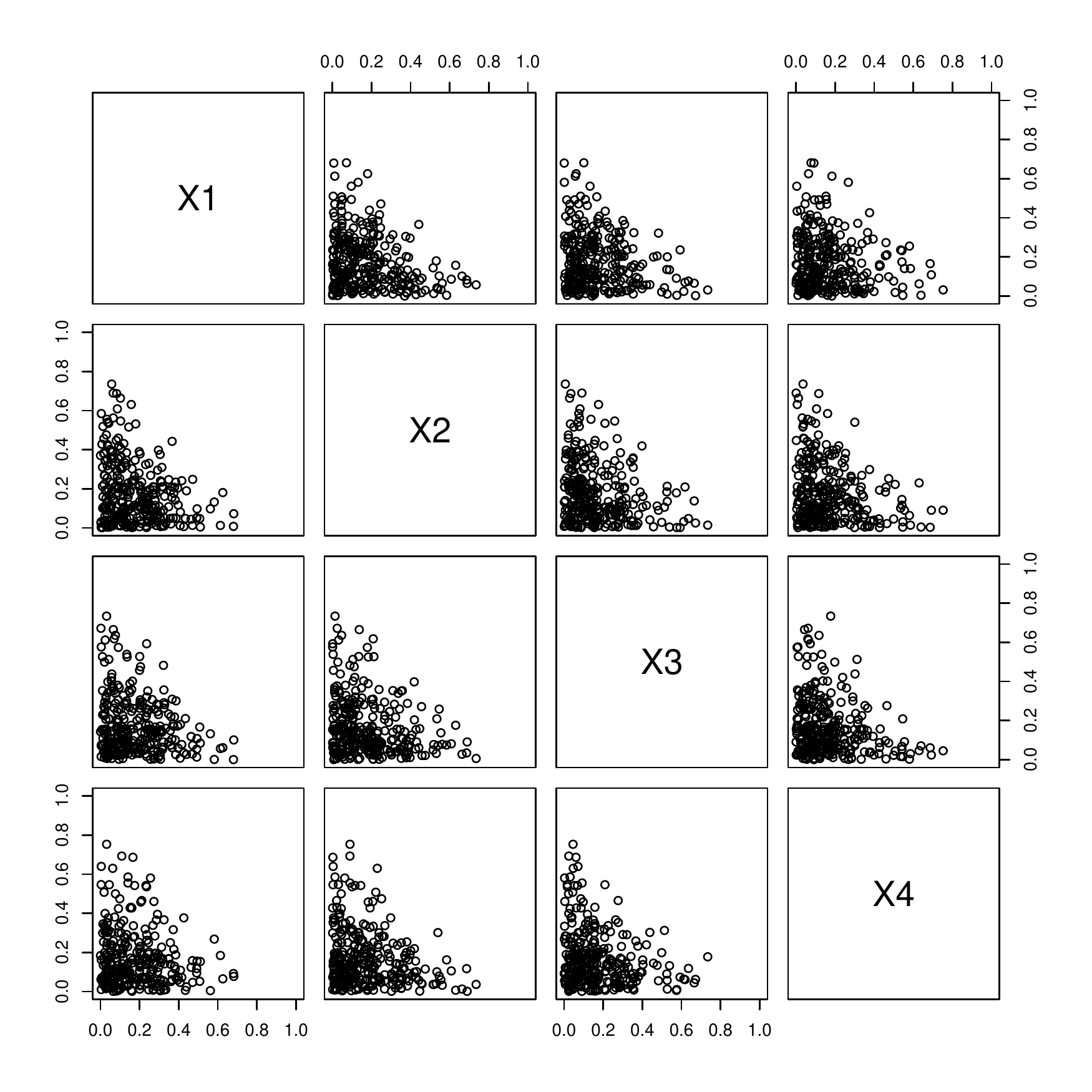}\includegraphics[width=0.25\hsize]{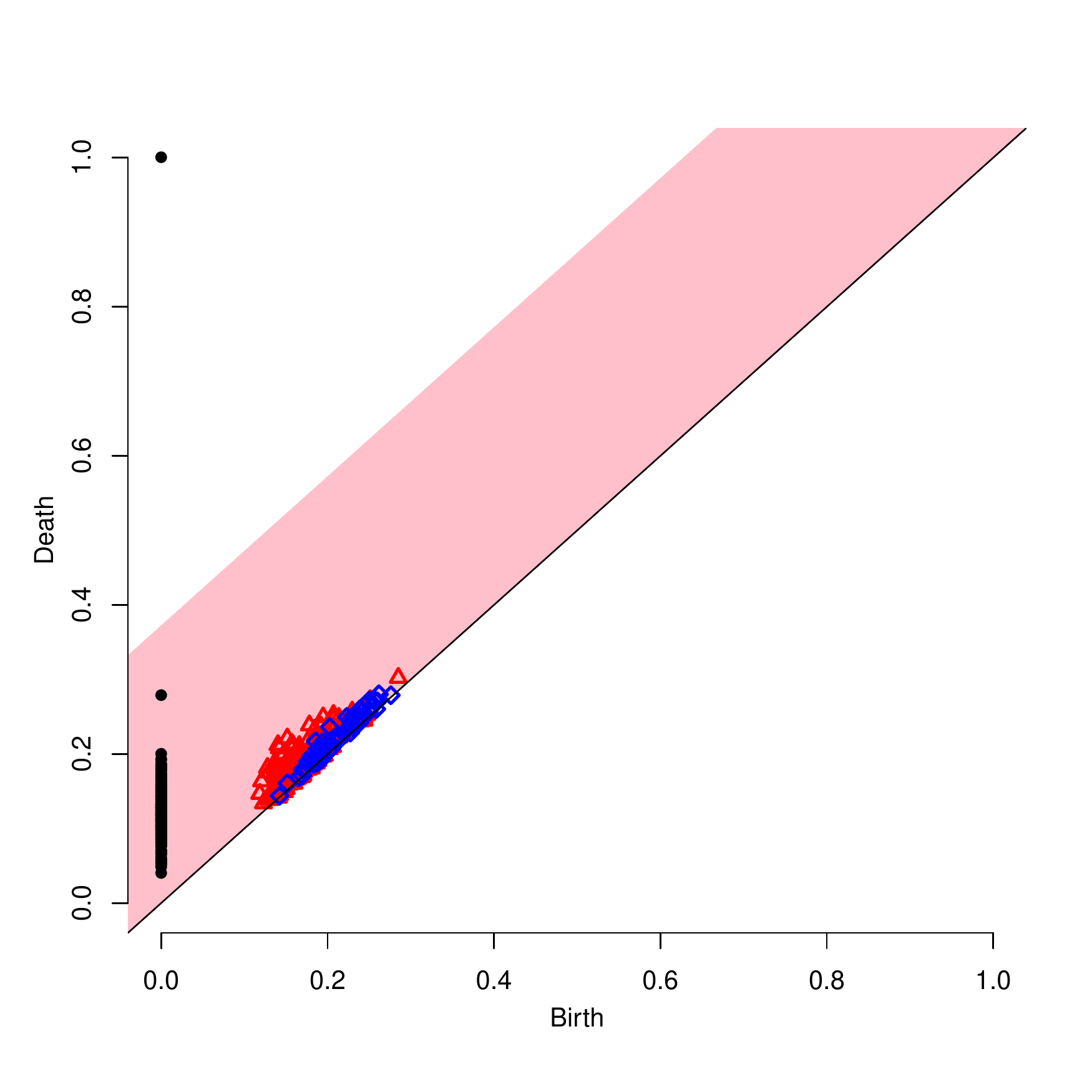}\label{fig:GMED}}\\
\subfloat[(12,3)-DTLZ5]{\includegraphics[width=0.25\hsize]{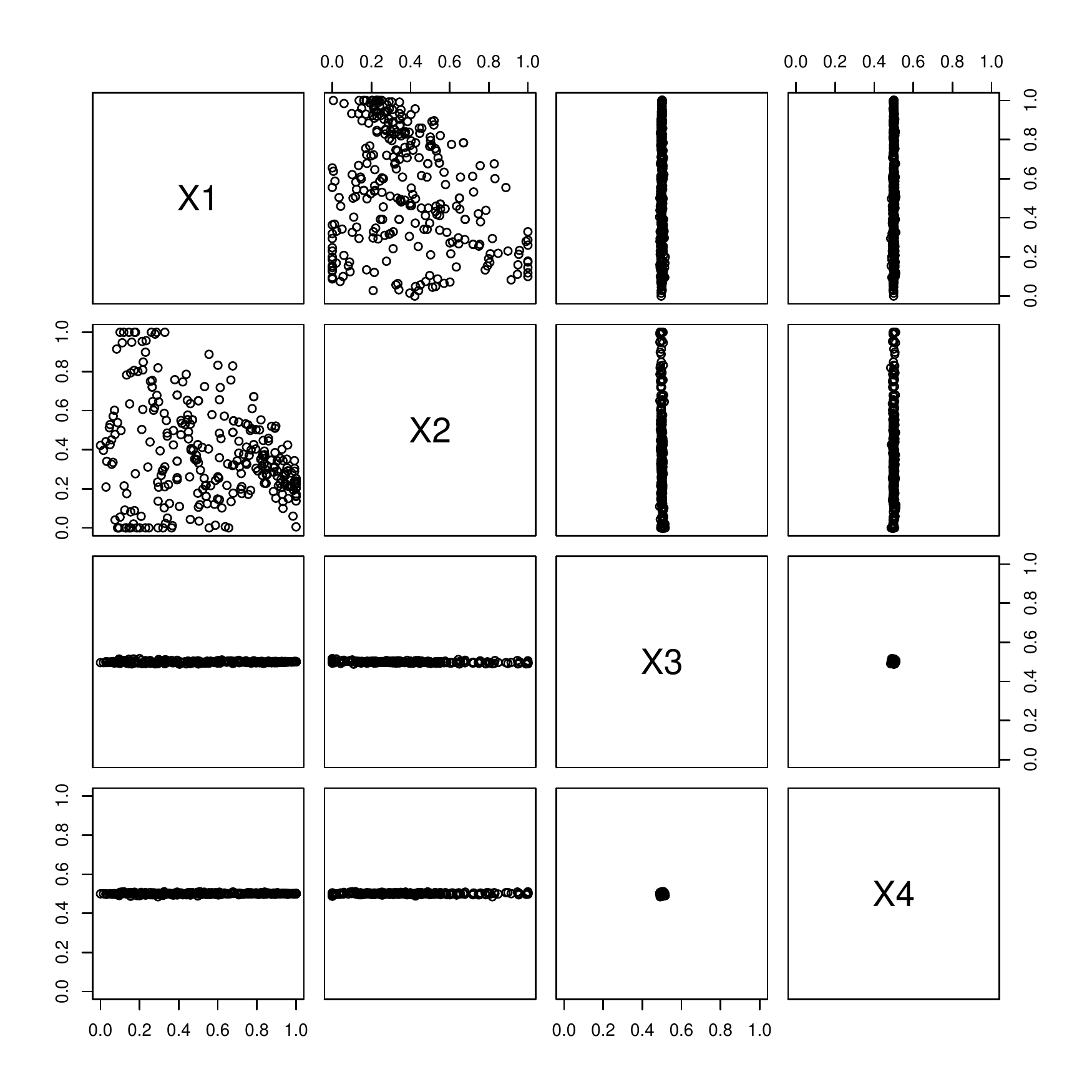}\includegraphics[width=0.25\hsize]{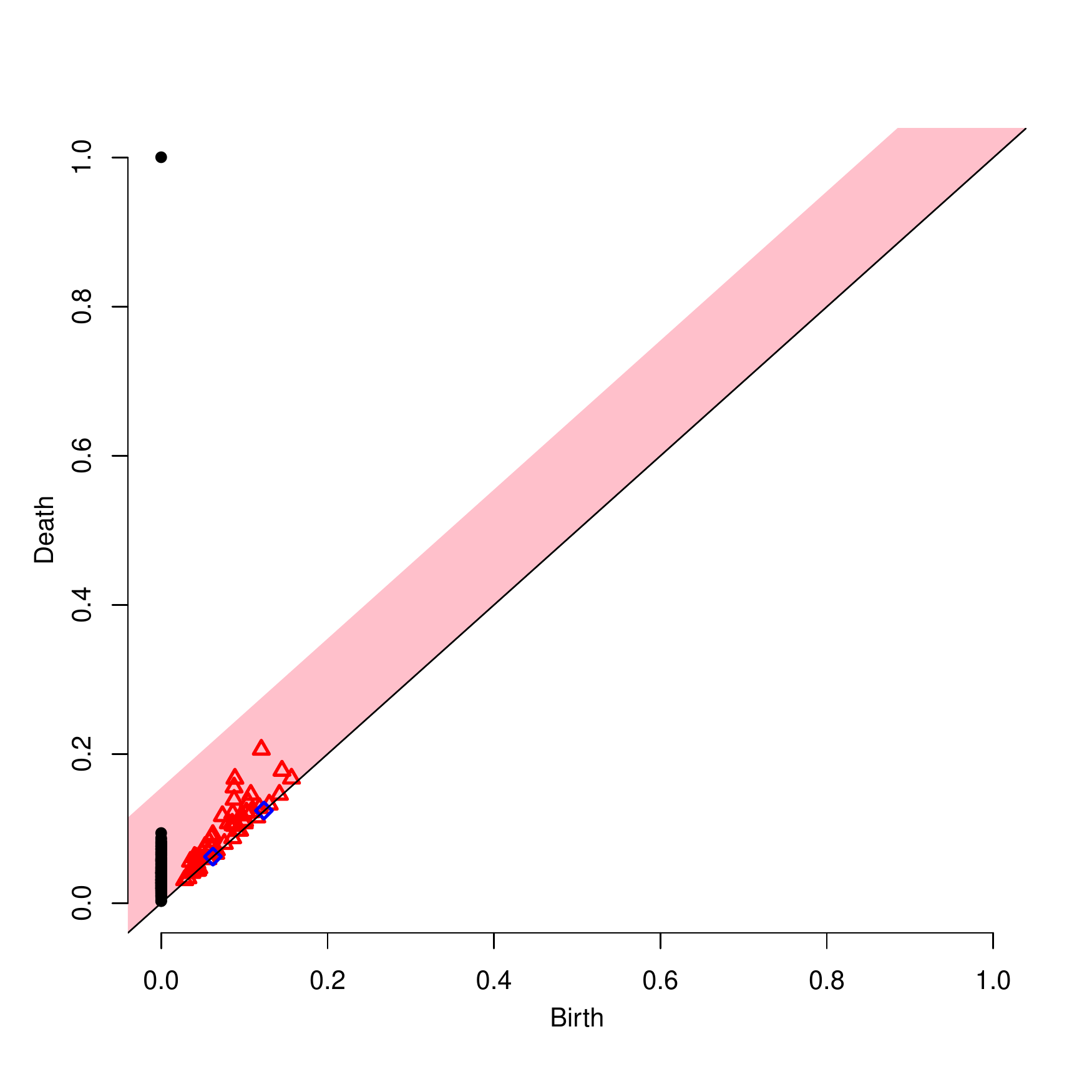}\label{fig:DTLZ5}}
\subfloat[(22,3)-DTLZ7]{\includegraphics[width=0.25\hsize]{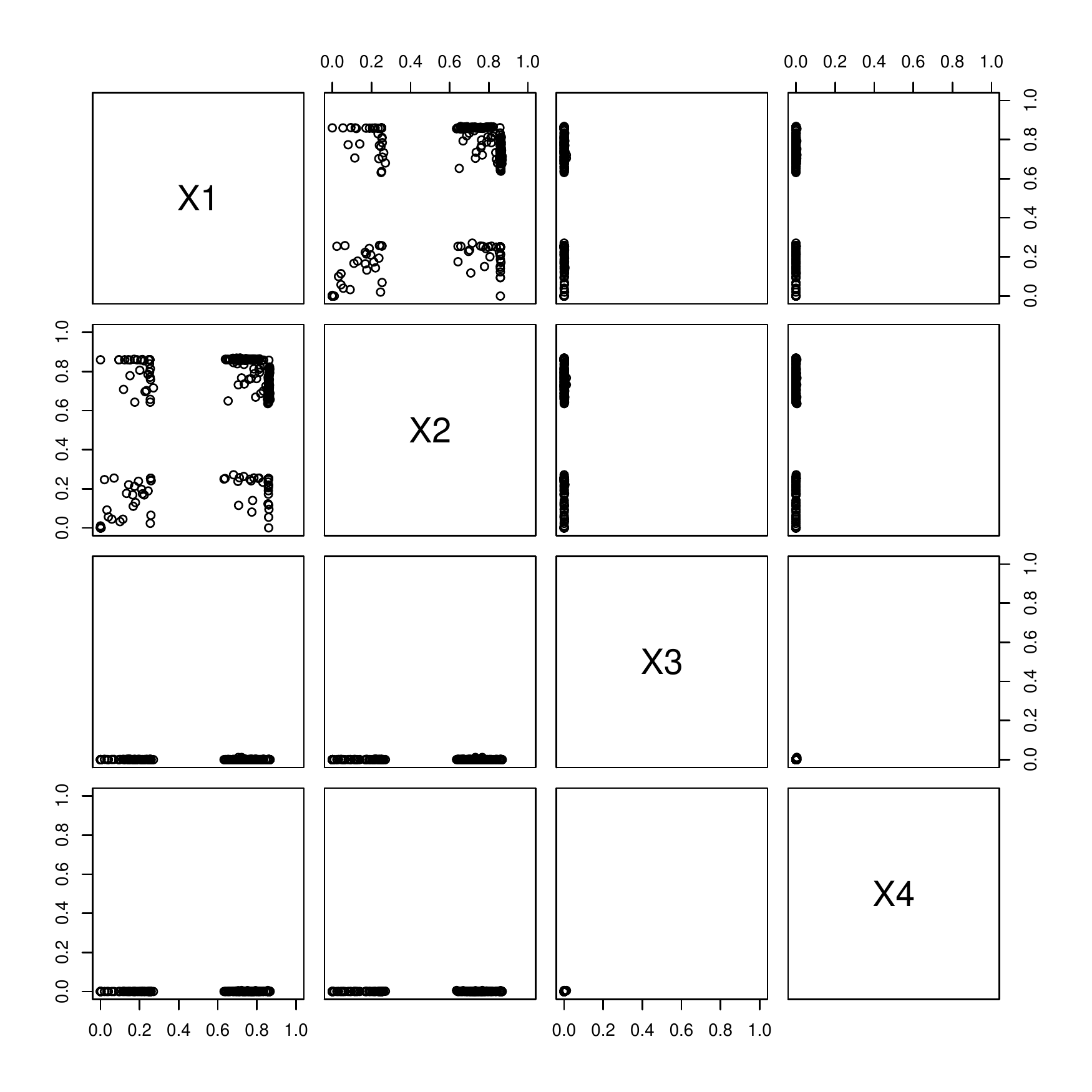}\includegraphics[width=0.25\hsize]{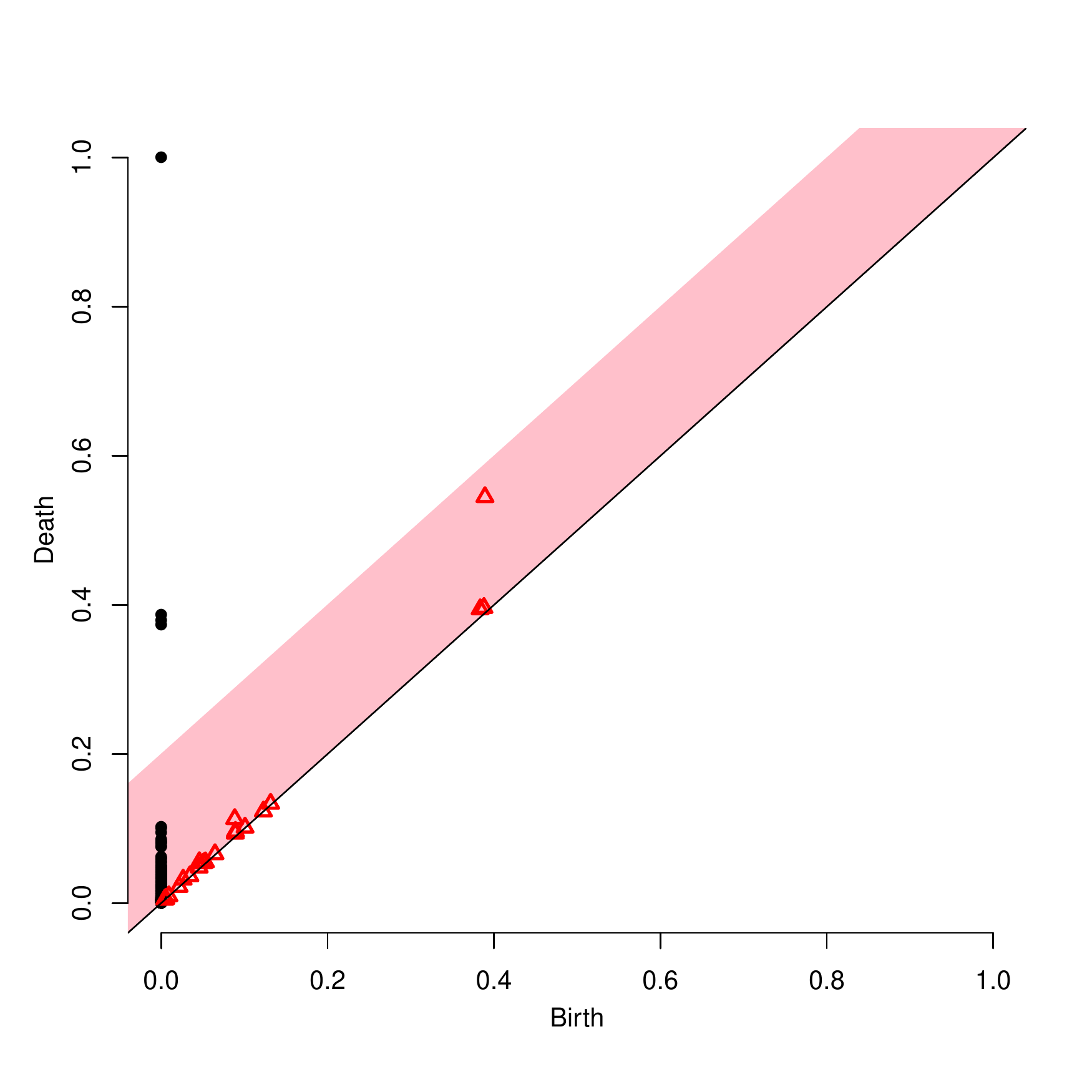}\label{fig:DTLZ7}}
\caption{Pareto set samples and their persistent diagrams of a typical trial.}\label{fig:plot}
\end{figure*}
The reason why our method failed to detect discontinuity is that numerical analysis in finite resolution cannot distinguish between ``discontinuous gaps'' and ``steep slopes'' of a function.
To detect discontinuity, we need to take a threshold of the function's Lipschitz constant and check whether each convex hull in \Cref{eq:simplicial-map} has a slope under the threshold or not.
To determine such a threshold, some techniques should be developed.

To understand what is going on, we pick a typical trial and show its sample and computed persistent diagram in~\Cref{fig:plot}.
There are many cycles in the confidence band and only one 0-cycle far out of the band.
Points near the diagonal line are to be noise while the point far from the diagonal is signal.

\begin{figure*}[t!]
\centering%
\subfloat[Wall time]{\includegraphics[width=0.33\hsize]{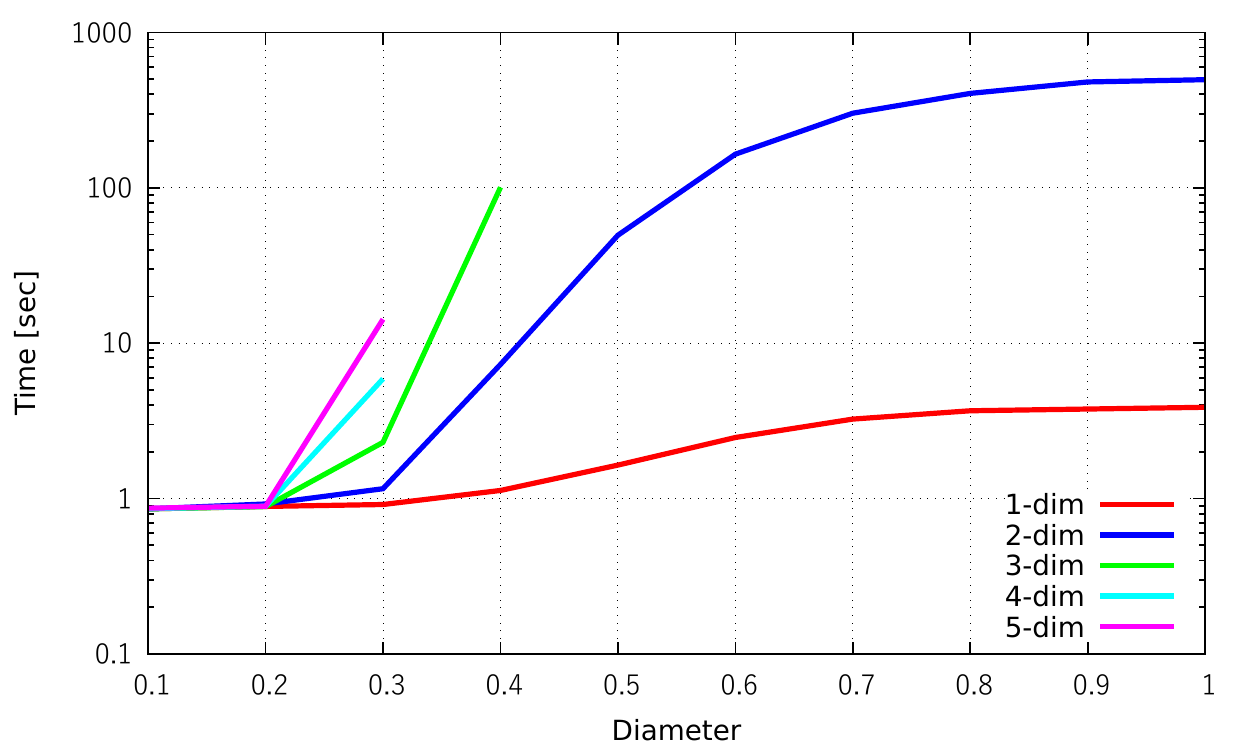}}
\subfloat[Peak memory usage]{\includegraphics[width=0.33\hsize]{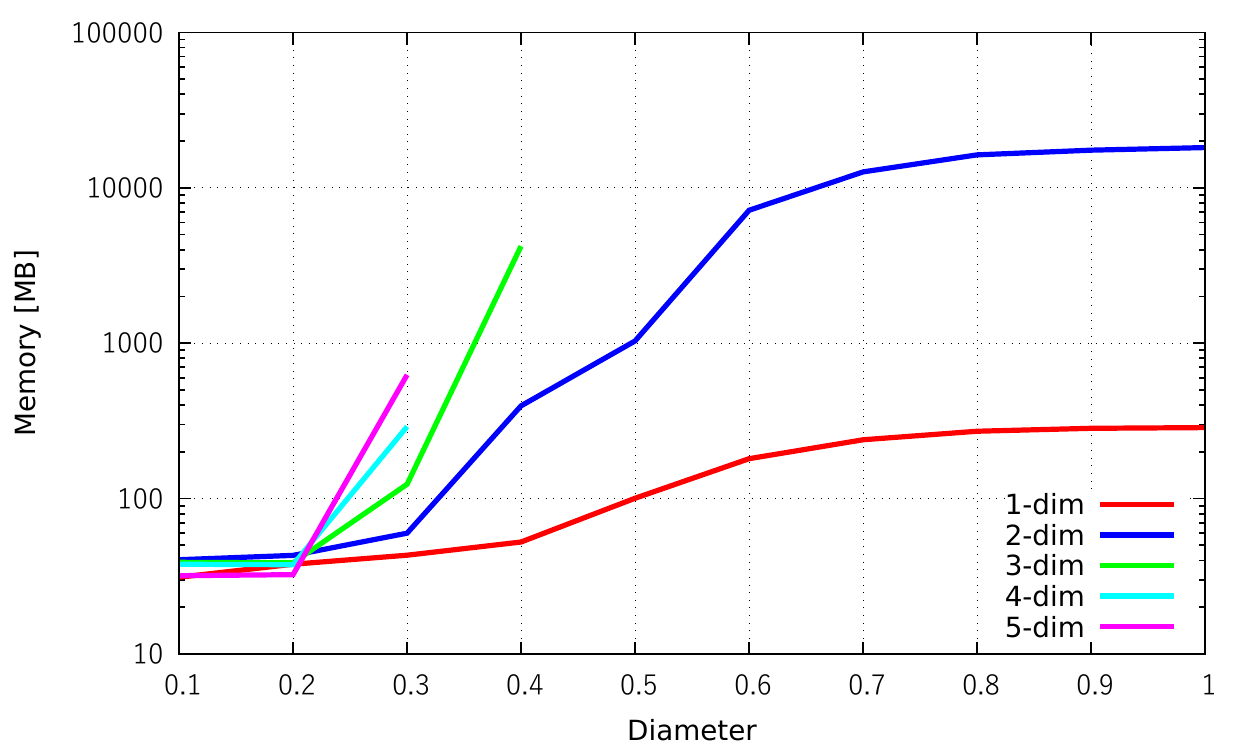}}
\subfloat[Number of simplices]{\includegraphics[width=0.33\hsize]{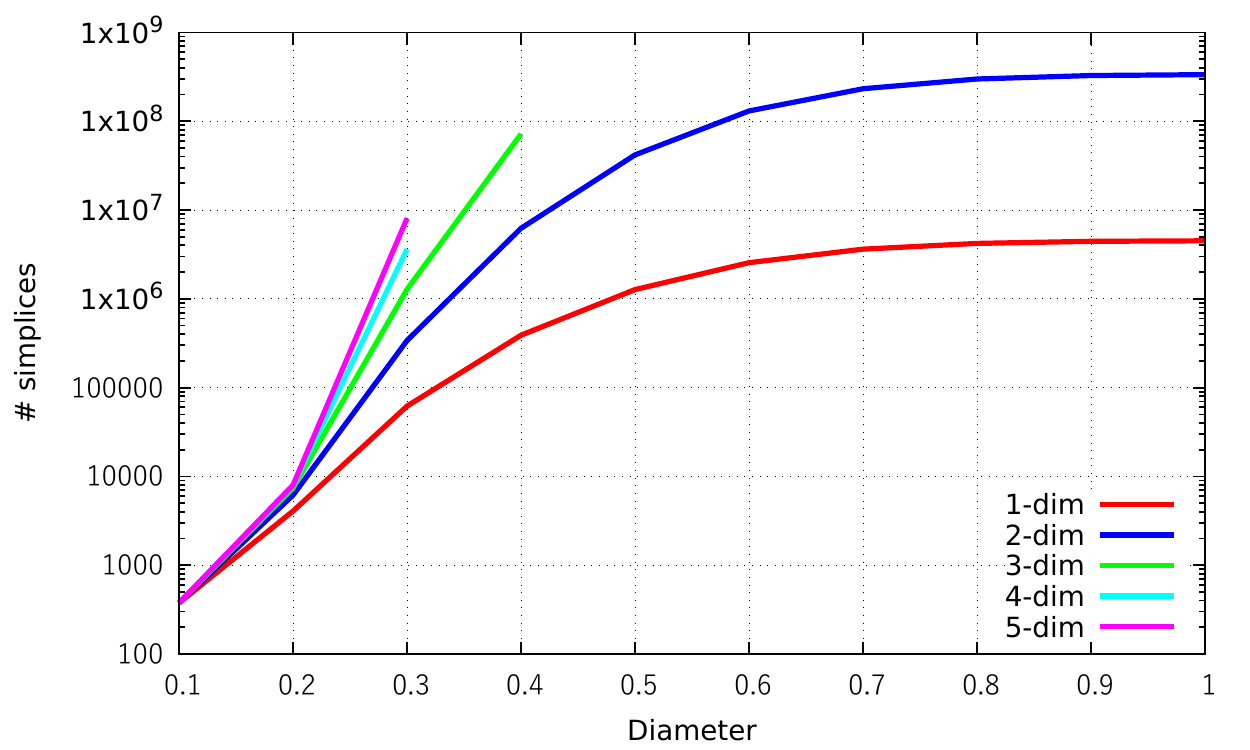}}
\caption{Computational costs for (40,6)-MED (averaged over 10 trials).}
\label{fig:costs}
\end{figure*}

The experiments in the previous section constructed a 2-Rips complex and computed up to 2-cycles.
Ideally, it is needed to check higher dimensions up to the dimensionality of the Pareto set arising in the problem.
A generic $n$-variable, $m$-objective problem has a $\min(n, m-1)$-dimensional Pareto set.
There can arise at most $\min(n-1, m-1)$-homology group\footnote{Since $\R^n$ can contain up to $(n-1)$-spheres.}.
We also conducted an experiment that tried to compute a higher-dimensional Rips complex but it was not finished as shown in~\Cref{fig:costs}.
We need an algorithm for building complexes that is scalable to higher dimensions.

\subsection{Computational Costs}
Ideally, we need to compute our algorithm for every subproblem rather than do only for the original problem.
The test for (S2) violation should check intersection between every pair of simplices, not for the highest dimension but for all dimensions.
However, the brute-force checking is impractical due to computational costs since an $m$-objective problem has $2^m - 1$ subproblems and a $k$-complex has $\Omega(2^k)$ simplices.
Smarter algorithms should be developed for complete computation.

\subsection{Related Approaches}
Since the persistent diagram has $O(\exp(N))$ time and space complexity with respect to sample size $N$, the use of large samples or computation of higher-degree homologies are prohibited.
To reduce the complexity, a subsampling technique~\cite{Chazal15} has been proposed recently.
This technique may partially resort the scalability issue for our method but still more effort would be needed.

There are several studies that develop techniques for estimating stratifications from data using intersection homology and local homology~\cite{Bendich07,Bendich10,Bendich11,Bendich12}.
These approaches do not assume domain specific knowledge and thus are general but require a larger sample than our approach.
Roughly speaking, those methods require thousands of points to recognize the topology of 2D or 3D objects.
By exploiting the properties of simple problems, our method relied on only 300 points to correctly estimate the true topological information, which seems reasonable for the optimization purpose.

\section{Conclusions}\label{sec:conclusions}
In this paper, we proposed a topological data analysis (TDA) technique to detect the non-simplicity of a given optimization problem.
As a part of our method, persistent homology is used to determine the simplex diameter of a Rips complex to recover the true topology of the Pareto set and the Pareto front.
We have confirmed via numerical experiments that the diameter determined by this method is appropriate to judge whether a given problem is simple or not.

Since TDA is a currently developing research area, existing algorithms for computing persistent homology suffer from the scalability issue shown in \Cref{fig:costs}.
We need smarter algorithms and data structures of complexes that are scalable to higher dimension or larger sample size.
In particular, reducing the complex size is an important research topic.

For reliable estimation, the proposed method requires a fine-grained approximation of the Pareto set: sample points must distribute over the true Pareto set, closely, uniformly and entirely.
Such a set is often unattainable in real-life applications.
Evaluating the robustness of the proposed method in more wild conditions is future work.

\bibliographystyle{ACM-Reference-Format}
\bibliography{ref}

\end{document}